%% file: main.tex
\documentclass[10pt,twocolumn,letterpaper]{article}

\usepackage{wacv}              

\usepackage{graphicx}
\usepackage{amsmath}
\usepackage{amssymb}
\usepackage{booktabs}

\usepackage{xcolor}
\usepackage{amsfonts}
\usepackage{multirow}
\usepackage{enumitem}
\usepackage{comment}
\usepackage{bbm}
\usepackage{arydshln} 
\usepackage{wrapfig}
\usepackage{subcaption}

\usepackage[pagebackref,breaklinks,colorlinks]{hyperref}

\usepackage[capitalize]{cleveref}
\crefname{section}{Sec.}{Secs.}
\Crefname{section}{Section}{Sections}
\Crefname{table}{Table}{Tables}
\crefname{table}{Tab.}{Tabs.}

\def\wacvPaperID{623} 
\def\confName{WACV}
\def\confYear{2024}

\begin{document}

\title{Tunable Hybrid Proposal Networks for the Open World}

\author{Matthew Inkawhich\textsuperscript{1}\quad Nathan Inkawhich\textsuperscript{2}\quad Hai Li\textsuperscript{1} \quad Yiran Chen\textsuperscript{1}\\
\textsuperscript{1}Duke University\quad \textsuperscript{2}Air Force Research Laboratory\\
{\tt\small matthew.inkawhich@duke.edu}
}
\maketitle

\begin{abstract}
    Current state-of-the-art object proposal networks are trained with a closed-world assumption, meaning they learn to only detect objects of the training classes. These models fail to provide high recall in open-world environments where important novel objects may be encountered. While a handful of recent works attempt to tackle this problem, they fail to consider that the optimal behavior of a proposal network can vary significantly depending on the data and application. Our goal is to provide a flexible proposal solution that can be easily tuned to suit a variety of open-world settings. To this end, we design a Tunable Hybrid Proposal Network (THPN) that leverages an adjustable hybrid architecture, a novel self-training procedure, and dynamic loss components to optimize the tradeoff between known and unknown object detection performance. To thoroughly evaluate our method, we devise several new challenges which invoke varying degrees of label bias by altering known class diversity and label count. We find that in every task, THPN easily outperforms existing baselines (e.g., RPN, OLN). Our method is also highly data efficient, surpassing baseline recall with a fraction of the labeled data.
\end{abstract}

\input{sections/introduction.tex}

\input{sections/related_work.tex}

\input{sections/learning_open_world_proposals.tex}
\input{sections/thpn}
\input{sections/experiments.tex}
\input{sections/conclusion.tex}

{\small
\bibliographystyle{ieee_fullname}
\bibliography{egbib}
}

\input{sections/appendix.tex}

\end{document}

%% file: sections/introduction.tex
\vspace{-2mm}
\section{Introduction}

The goal of object proposal generation is to detect and localize all potential objects of interest in an image. High-quality object proposals serve as the foundation for many vision-based applications including object detection \cite{DeepBox, RCNN, FastRCNN, FasterRCNN, CascadeRCNN}, segmentation \cite{semantic_segmentation_using_regions_and_parts, conv_feature_masking, MaskRCNN}, object discovery \cite{object_discovery_in_the_wild, UnsupervisedJointObjDiscovery, DissimilarityCoefficientWeaklySupervised}, and visual tracking \cite{UnsupervisedObjectDiscovery, RobustVisualTracking}. Over recent years, heuristic-based object proposal algorithms \cite{SelectiveSearch, EdgeBoxes, MCG} have been supplanted by deep learning-based solutions such as Region Proposal Network (RPN) \cite{FasterRCNN} which provide superior recall and faster inference. Currently, there is a significant push towards creating models that can function in open-set \cite{OGOpenSet, ProbModelsOpenSet, OverlookedElephantOpenSet} and open-world \cite{OGOpenWorld, TowardsOpenWorldObjectDetection} environments. Here, the deployed model will encounter known object classes from the labeled training distribution as well as novel classes. We refer to these instances as ``in-distribution'' (ID) and ``out-of-distribution'' (OOD) objects, respectively. An ideal object proposal model would detect all ID \textit{and} OOD objects of interest with high confidence. However, most existing proposal networks overfit to the ID classes, meaning that if we deploy them in an open-world setting many OOD objects will go undetected \cite{OverlookedElephantOpenSet}. In a real-world system this kind of mistake could have catastrophic consequences. 
While several recent works improve a classifier's ability to discern ID from OOD objects \cite{MSP, OutlierExposure, EnergyBasedOOD, NateOOD, JingyangOOD, VOS}, we argue that the \textit{proposal network} is holding back open-set/world detection. Ultimately, if an OOD object is not confidently proposed, the region will never even reach the classification stage.

\begin{figure}[t]
  \centering
   \includegraphics[width=0.95\linewidth]{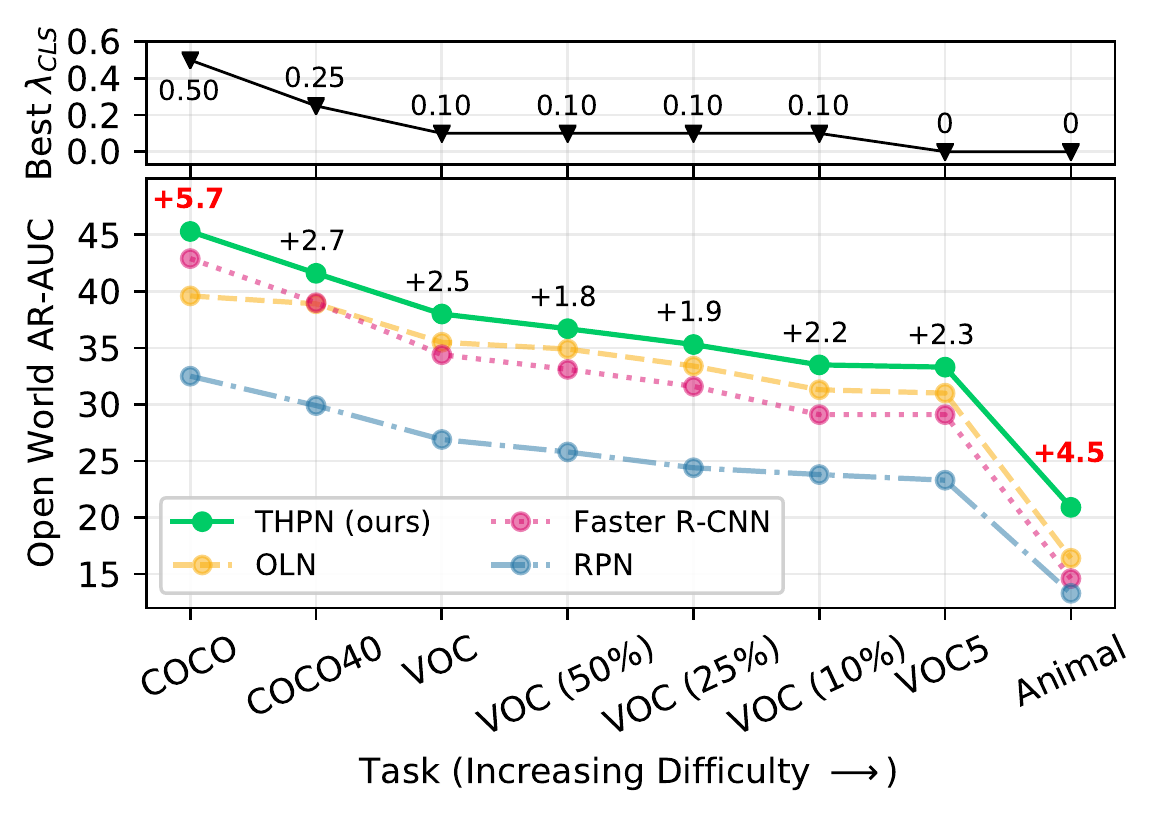}
   \vspace{-3mm}
   \caption{THPN's adjustability allows it to excel across a broad range of open world tasks. \textit{Top:} optimal $\lambda_{CLS}$ for each task. \textit{Bottom:} recall comparison; annotations are THPN's margins over OLN \cite{OLN}.}
   \label{fig:summary}
   \vspace{-3mm}
\end{figure}

The primary motivation for this work is to design a proposal network that is useful in a variety of real-world settings. To do this, we must expand the settings that we use to evaluate the models. Currently, the litmus test for open-set/world object detection performance involves training on one diverse natural imagery dataset and testing it on another (e.g., train on PASCAL VOC \cite{PASCALVOC}, test on COCO \cite{COCO}) \cite{OverlookedElephantOpenSet, OLN, KevinsPaper, LearningToDetectEveryThing}. While this style of evaluation is convenient, it emulates a mere sliver of potential open-world scenarios that we may encounter in the real-world. Existing evaluations make two key assumptions: (1) they assume we have access to a large diverse set of training data that contains exhaustive labels for nearly every class that we would want to detect during deployment; and (2) they assume that all open-world applications consider \textit{every} unlabeled object to be of interest. In reality, both the quality of the labeled data and the desired behavior of the model can vary significantly depending on the application in which it is used. 
For example, in robotics applications it may be more important to localize all potential regions of interest, down to the level of ambient objects such as light switches and power outlets. However, in an application such as a vehicle identification system, it is more critical to detect novel types of vehicles than, for instance, buildings and trees. In this work, we design several novel challenges to simulate varying degrees of label bias to more rigorously evaluate our method. Specifically, a \textit{training class diversity} challenge restricts ID class coverage, a \textit{semi-supervised} challenge directly reduces the amount of labeled samples we have, and a \textit{ships} challenge tests the models in a different domain with a uniquely constrained set of OOD objects of interest.

To address these challenges, we develop a Tunable Hybrid Proposal Network (THPN) that leverages two types of object representation: (1) classification-based objectness and (2) localization-based objectness. Classification-based objectness is employed in the canonical Region Proposal Network (RPN) \cite{FasterRCNN, CascadeRPN, RegionProposalGuidedAnchoring}, and frames object learning as a discriminatory task. This works well for detecting ID objects, but struggles to detect OOD objects as it explicitly learns that all non-labeled regions are \textit{background} \cite{OLN, FSODwoutForgetting, KevinsPaper}. Localization-based objectness, introduced recently by Kim et al.'s Object Localization Network (OLN) \cite{OLN}, frames objectness as the localization quality \cite{FCOS, IoUNet} between a region and any ground truth box. This approach promotes a less discriminative detector that more readily generalizes to dissimilar OOD classes. By using both representations simultaneously, THPN is capable of the best of both worlds. The behavior of THPN can be easily tuned with a single hyperparameter $\lambda_{CLS} \in [0,1]$ which determines how significantly the model weights classification-based objectness versus localization-based objectness. Depending on the needs of the application, THPN can operate as a conservative ID expert using a large $\lambda_{CLS}$, an aggressive OOD object detector using a small $\lambda_{CLS}$, or anywhere in between. In addition, THPN uses a novel open-world-aware self-training procedure which bolsters the existing label set with high-quality pseudo-labels \cite{Lee_pseudo-label}. Unlike common self-training solutions \cite{Lee_pseudo-label, SelfTrainingSurvey, SimpleSSLFrameworkforObjDet}, our approach does not require any auxiliary samples to generate pseudo-labels on, and does not require full retraining in each round. Finally, we develop a dynamic loss to address challenges such as class-imbalance and imperfect pseudo-label targets.

THPN outperforms all baselines in all evaluation settings that we consider. 
On the common VOC$\rightarrow$COCO open-set benchmark, where models are trained on VOC-class labels and tested on non-VOC COCO classes, THPN vastly improves upon RPN (+18.9\% AR100) and OLN (+5.7\% AR100).
\cref{fig:summary} shows a summary of results across several of our novel evaluation challenges in terms of ALL object recall. Note that THPN can easily surpass OLN in more difficult biased tasks without sacrificing performance on low-bias tasks.
For example, THPN trained on a five-class subset of VOC achieves higher OOD recall than an OLN trained on the entire 20-class VOC subset. Furthermore, a THPN trained on a random 10\% subset of the original VOC labels rivals the OOD recall of an OLN trained on 100\% of the labels. 
On the \textit{ships} challenge, THPN achieve a $\sim$3x recall improvement over Faster R-CNN on OOD ships.
Overall, THPN's flexibility enables it to be a better general solution for open-set/world detection problems.

%% file: sections/related_work.tex
\begin{figure*}
  \centering
  \begin{subfigure}[t]{0.35\textwidth}
    \includegraphics[height=1.6in]{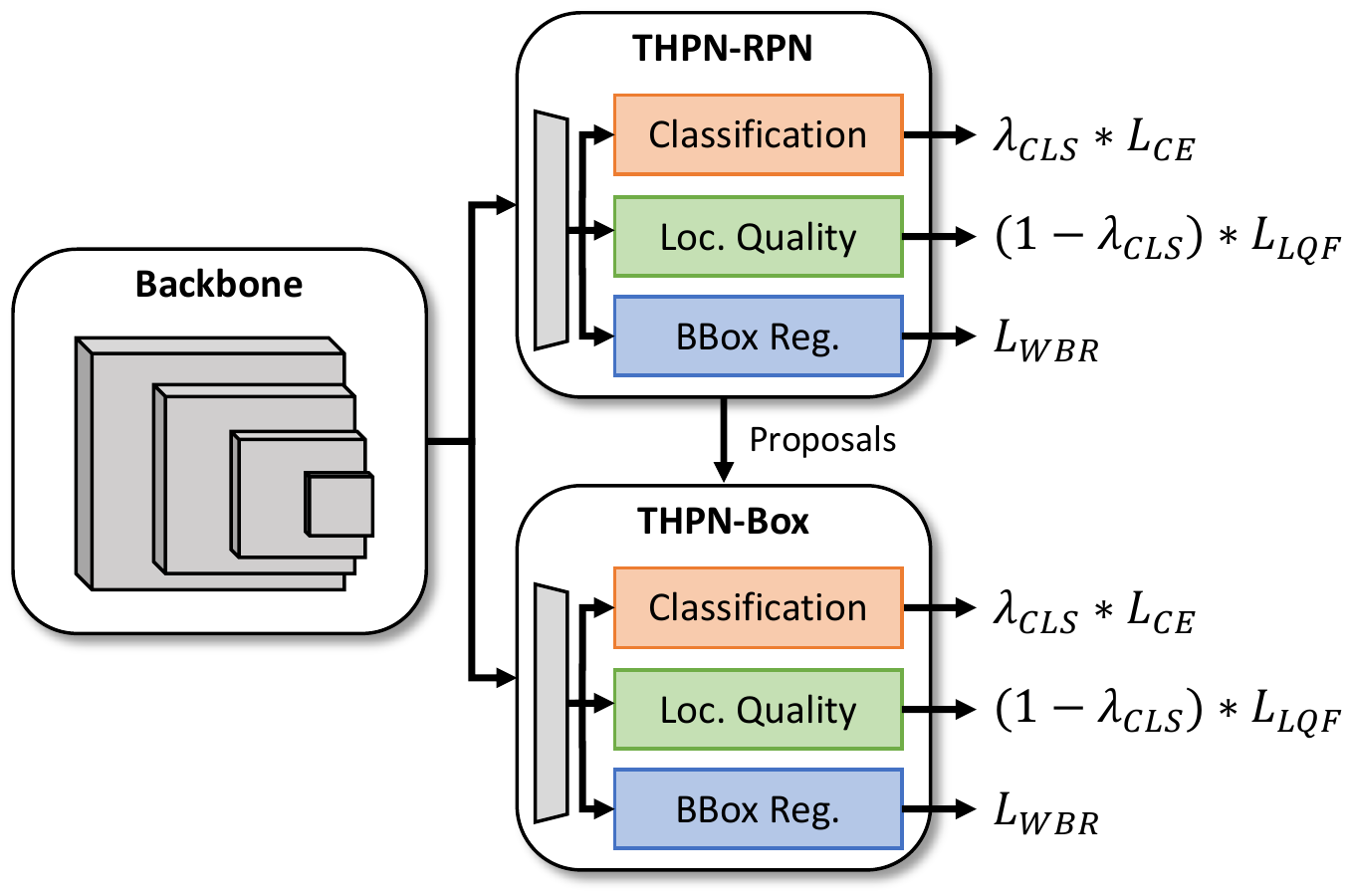}
    \caption{THPN's hybrid architecture has heads for both classification-based and localization-based objectness. The total objectness loss is a linear interpolation between the $CLS$ and $LQ$ head losses.}
    \label{fig:THPN_architecture}
  \end{subfigure}
  \hfill
  \begin{subfigure}[t]{0.62\textwidth}
    \includegraphics[height=1.8in]{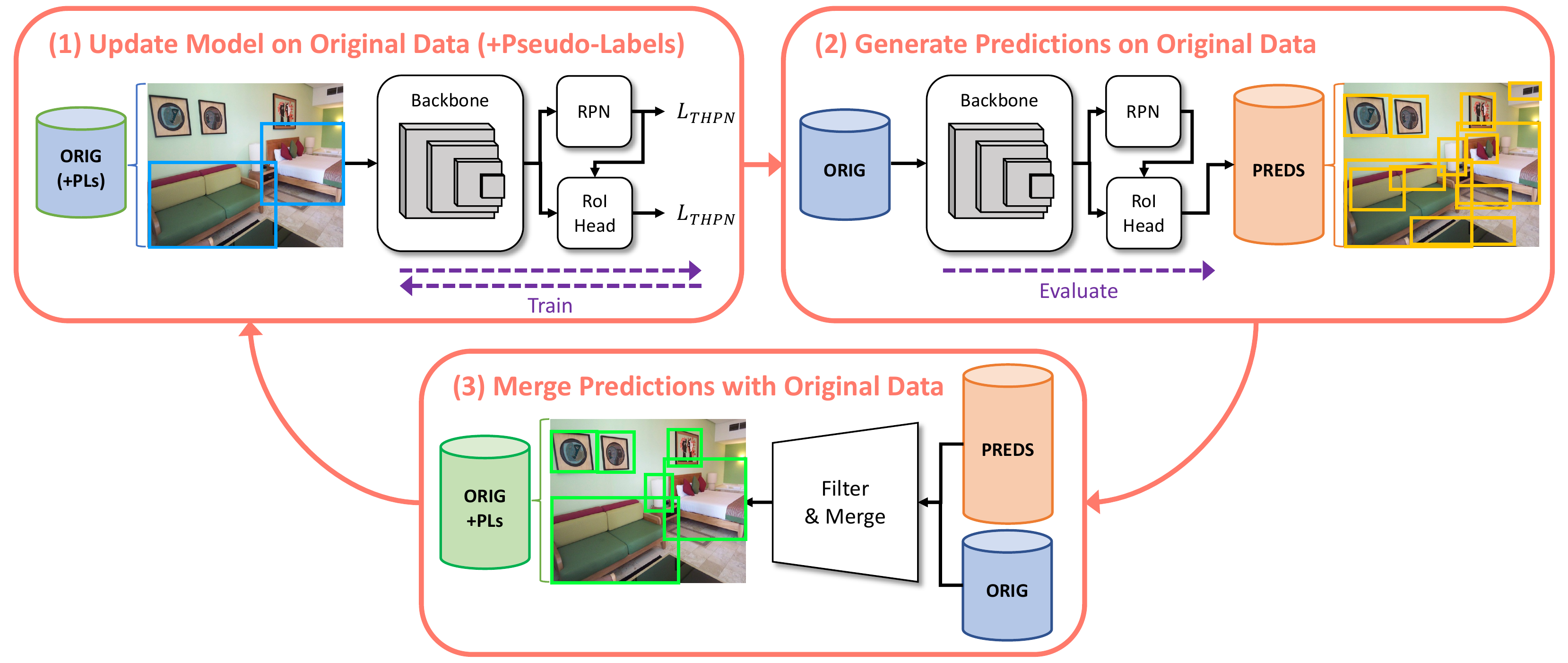}
    \caption{Our self-training algorithm consists of three stages. First, update the model on the original data (plus pseudo-labels we may have). Second, use the trained model to generate predictions on the \textit{original} images. Third, filter and merge the predictions with the original ground truth labels to create an updated label set with pseudo-labels (PLs). Repeat this process iteratively.}
    \label{fig:self-training}
  \end{subfigure}
  \vspace{-2mm}
  \caption{Overview of THPN's architecture and training procedure.}
  \vspace{-3mm}
  \label{fig:overview}
\end{figure*}

\section{Related work}
\textbf{Class-agnostic object proposal.~}
Early methods for class-agnostic object detection rely on handcrafted image features such as Gaussian filters and edges \cite{MeasuringObjectnessofImageWindows, GeodesicObjectProposals, SelectiveSearch, EdgeBoxes, MCG}, but the advent of deep learning has rendered these heuristic-driven approaches obsolete \cite{DeepBox, FasterRCNN}. RPN and its variants \cite{FasterRCNN, CascadeRPN, RegionProposalGuidedAnchoring} \textit{learn} to identify a reduced set of regions that have a high likelihood of containing objects. RPN can be trained inline as part of a two-stage detection architecture \cite{FasterRCNN, MaskRCNN, CascadeRCNN, FPN} to attain impressive accuracy on ID classes. The problem with RPN is that it overfits to the ID categories \cite{OLN, KevinsPaper, LearningToDetectEveryThing}. Object Localization Network (OLN) \cite{OLN} combats this overfitting by replacing the classification heads of a class-agnostic Faster R-CNN with localization quality prediction heads to avoid treating OOD objects as \textit{background}. 
Konan et al. \cite{KevinsPaper} use unknown object masking to reduce the number of false-negative regions sampled during training. Finally, Saito et al. \cite{LearningToDetectEveryThing} use a background erasing augmentation and a multi-domain training strategy to reduce the bias of classification-based proposal networks.
Uniquely, our solution combines both objectness representations with a novel self-training procedure to better address a variety of open-world scenarios. 

\textbf{Open-set/world detection.~}
Unlike class-agnostic proposal networks, full object detection models also classify the objects. \textit{Open-set} detectors accept that OOD objects will inevitably be encountered during inference, and attempt to flag them as \textit{unknown}.
Dhamija et al. \cite{OverlookedElephantOpenSet} find that closed-set models frequently misclassify OOD objects as ID classes despite training with an explicit \textit{background} class. Miller et al. \cite{Miller1, Miller2} use dropout sampling \cite{DropoutSampling} to estimate uncertainty and reduce these open-set false positives. Recently, virtual outliers \cite{VOS} and contrastive learning \cite{ExpandingLowDensityLatentRegionsOSOD, SupervisedContrastiveLearning}, have been used to separate known and unknown instances in feature space. Joseph et al. \cite{TowardsOpenWorldObjectDetection} present the first attempt at an \textit{open-world} detection system, which not only detects OOD objects, but also incrementally learns the newly encountered classes on the fly. Since then, several works have incrementally improved open-world detection \cite{RevisitingOWOD, OWDETR, UCOWOD, OWODDiscriminativeClassPrototype}. Critically, these open-set/world detectors rely on a classification-based RPN to provide proposals for both ID and OOD objects, meaning that many OOD objects are likely to go undetected. In this work, we focus on the development of a more powerful proposal network, which can be integrated into such systems in the future.

\textbf{Self-training.~}
Self-training is a powerful technique in semi-supervised learning, where only a subset of the training data has labels \cite{SelfTrainingSurvey, OverviewDeepSSL}. Based on the classic concept of a delay-feedback algorithm \cite{Scudder65a, Fralick67}, self-training uses a trained model to generate pseudo-labels on unlabeled data which are then used to bolster the existing training set, leading to better subsequent models \cite{Lee_pseudo-label, SelfTrainingSurvey, RethinkingPretrainingSelftraining}. While self-training is most commonly employed in image classification tasks \cite{BillionScale, SelfTrainingWNoisyStudent, CReST}, some works have used it for the object detection task \cite{RosenbergHS05, RethinkingPretrainingSelftraining, SimpleSSLFrameworkforObjDet, InteractiveSelfTrainingMeanTeachers, LabelVerifyCorrect} to improve closed-set performance in semi-supervised and few-shot scenarios. 
Different from previous works, we apply the principles of self-training to the open-world proposal problem. We find that this setting allows us to loosen many of the requirements made by existing approaches \cite{Lee_pseudo-label, OverviewDeepSSL, SimpleSSLFrameworkforObjDet}. For example, our method requires no auxiliary data, and iteratively fine-tunes the same model rather than fully re-training it. 

%% file: sections/learning_open_world_proposals.tex
\section{Learning open-set proposals}
\label{sec:learning_open_world_proposals}
To build intuition, we formalize the open-set object proposal problem. Generally, in an object detection task we have a set of \textit{known} (ID) object classes $\mathcal{K} = \{1, 2, \dots, C\} \subset \mathbb{N}^+$ that we have labels for. Typically there are also a considerable number of unlabeled instances of \textit{unknown} (OOD) classes $\mathcal{U} = \{C+1, \dots \} \subset \mathbb{N}^+$ that coexist with the known instances in the images. 
The goal of the open-set object proposal task is to train a model $\mathcal{M}$ parameterized by $\theta$ to detect and localize \textit{all} object instances of potential interest in a test set (i.e., all instances in the set $\mathcal{K} \cup \mathcal{U}$). For a given test image $X$, the proposal network's function is $\mathcal{M}(X; \theta) = \{[x,y,w,h,s]_{j=1\dots N}\}$, where $x$, $y$, $w$, and $h$ denote the center coordinates, width, and height of the bounding box, respectively. The predicted ``objectness" score $s \in [0,1]$ is the confidence that box $j$ contains an object. 
Although the proposal task differs from the full open-set detection task (in which the model also predicts the class of each object), most current state-of-the-art open-set/world detection systems rely on proposal networks to produce high-recall candidate regions \cite{OverlookedElephantOpenSet, ExpandingLowDensityLatentRegionsOSOD, TowardsOpenWorldObjectDetection, RevisitingOWOD}. Ultimately, the upper bound of performance achievable by such systems is premised on the recall of the proposal network.

%% file: sections/thpn.tex
\section{Tunable Hybrid Proposal Network (THPN)}

Our primary goal with THPN is to introduce a flexible proposal network that can be readily adapted to many open-world environments. 
Controllable by a single hyperparameter, our idea is to allow the user to adjust the model's willingness to detect OOD objects that are dissimilar to the labeled classes depending on their application's requirements 
To achieve this, we develop a novel training algorithm (\cref{sub:self-training_precedure}), model architecture (\cref{sub:model_architecture}), and dynamic loss (\cref{sub:model_architecture}). \cref{sub:implementation_details} contains implementation details.


\subsection{Self-training procedure}
\label{sub:self-training_precedure}
One major drawback of existing proposal networks is that their generalization is largely dependent on the quantity and diversity of the labeled training data. Self-training can significantly mitigate this issue by artificially adding labels to the dataset. Self-training is the process of training a model on available labeled data, running inference on unlabeled inputs to generate high-quality pseudo-labels, and training a new model on the union of the original training data and the pseudo-labeled set \cite{Lee_pseudo-label}. While self-training is most commonly used for semi-supervised learning of closed-set tasks \cite{Lee_pseudo-label, OverviewDeepSSL, SimpleSSLFrameworkforObjDet}, we are the first to tailor this powerful regularization for open-world object proposals. 
Specifically, we develop a three-stage self-training algorithm that is outlined in \cref{fig:self-training}. The overall workflow is as follows. In Stage 1, we train the model on the original labeled data; in Stage 2, we evaluate the trained model \textit{on the original training images} to generate predictions; and in Stage 3, we filter predictions by score and merge the highest scoring predictions with the original ground truth labels. 
We can repeat this loop by training the model again on the \textit{updated} label set to incrementally improve pseudo-label quality and thus subsequent model generalization. 

Note that unlike existing self-training implementations, our method does not require the user to cull auxiliary unlabeled data. This is because in virtually all real-world detection data there exists a multitude of unlabeled OOD (and ID) objects that coexist in the same images as the labeled ID instances. Also note that existing self-training work \cite{Lee_pseudo-label, OverviewDeepSSL, SimpleSSLFrameworkforObjDet} retrains the model ``from scratch'' in each round. This approach is very expensive as it involves training for $(r*E)$ epochs, where $r$ is the number of self-training rounds and $E$ is the number of epochs in the standard training schedule. A more efficient approach is to repeatedly fine-tune the same model with the updated label set. We observe fine-tuning convergence within $E / 4$ epochs, so the total training cost of THPN is $(E + r*(E/4))$ epochs. 

Another important design detail is how we ``filter \& merge'' newly proposed boxes into the ground-truth label set in Stage 3. First, to avoid adding redundant labels we discard all predictions that overlap a ground truth box by 0.7 IoU. Next, we filter the remaining labels based on predicted objectness. While previous methods use simple thresholding of confidence \cite{Lee_pseudo-label, SimpleSSLFrameworkforObjDet, InteractiveSelfTrainingMeanTeachers}, we find that this approach does not provide enough granular control over the amount of predictions we allow to become pseudo-labels because DNNs are notoriously poorly calibrated \cite{on_calibration_of_dnns}. Instead, we take the top $P$ non-overlapping predictions, where $P$ is $p$\% of the number of original training instances ($p$ is a hyperparameter). With this approach, we can precisely control the amount of pseudo-labels we add relative to the number of original ground truth labels, making performance consistent regardless of the dataset size or the objectness metric used.

\subsection{Model architecture and losses}
\label{sub:model_architecture}
There are two known meta-strategies for learning-based proposal networks which are differentiated by how a region's ``objectness'' is quantified. \textit{Classification-based} approaches, such as RPN and class-agnostic Faster R-CNN \cite{FasterRCNN, TowardsOpenWorldObjectDetection}, directly predict a region's likelihood of containing an ID object. 
These models are trained to explicitly discriminate ID objects vs. background, meaning any OOD objects present in the training images are learned as negatives (i.e., background). 
Thus, these models significantly overfit to the training classes \cite{OLN, FSODwoutForgetting, KevinsPaper}. Alternatively, \textit{localization-based} (i.e., classification-free) models \cite{OLN} predict a region's localization quality (e.g., centerness \cite{FCOS}, IoU \cite{IoUNet}) with respect to the nearest ground-truth box and treat this as a notion of objectness. In essence, this changes the task from ``\textit{What is the likelihood this region contains an object?}'' to ``\textit{How well does this region localize the nearest object?}''. Because predicting localization quality is not discriminatory, the model is not explicitly biased towards the ID classes. 
While this allows for better OOD detection, it comes at the cost of reduced ID proficiency.
For more details on these methods, see Appendix A.

The key insight of our work is that the best objectness representation to use is dependent on the data and the desired behavior of the system. 
For example, applications that prioritize ID recall would benefit from classification-based objectness, while applications that require detecting \textit{all} objects would benefit from localization-based objectness. 
Our solution is to leverage a hybrid objectness representation that can be readily tuned to suit the full spectrum of applications and environments.
To realize this design we use a two-stage detection architecture,
where a first stage THPN-RPN produces a set of reasonable candidate regions, and a second stage THPN-Box refines these candidate regions and makes the final objectness prediction. To allow THPN to use both objectness representations, we use three prediction heads in THPN-RPN and THPN-Box (see \cref{fig:THPN_architecture}). 
For each anchor, a classification head ($CLS$) predicts the likelihood that a region contains an object, a localization quality head ($LQ$) predicts a quality score (i.e., centerness \cite{FCOS} in THPN-RPN and IoU \cite{IoUNet} in THPN-Box), and a bounding box regression head ($BOX$) predicts the box offsets.


The loss function for both THPN stages is defined as:
\begin{equation} \label{eq:THPN_loss}
\begin{split}
L_{THPN}\big(\{c_i\}, \{q_j\}, \{t_i\}\big) = \\
\lambda_{CLS} \frac{1}{N_{CLS}} \sum_i L_{CE}(c_i, c_i^*) \\
+ (1 - \lambda_{CLS}) \frac{1}{N_{LQ}} \sum_j L_{LQF}(q_j, q_j^*) \\
+ \lambda_{BOX} \frac{1}{N_{BOX}} \sum_i c_i^* L_{WBR}(t_i, t_i^*).
\end{split}
\end{equation}
Due to the fact that we use two different sets of sampled anchors (based on different sampling criteria) to compute the losses, we use $i$ to denote the indexes of anchors for the $CLS$ and $BOX$ heads, and $j$ to denote the indexes of anchors for the $LQ$ head. Thus, $c_i$, $q_j$, and $t_i$ are the predicted object likelihood, localization quality score, and box offsets, respectively, and $c_i^*$, $q_j^*$ and $t_i^*$ are the corresponding targets. 
The total loss is composed of three terms. The first is the cross-entropy loss $L_{CE}$ from the $CLS$ head; the second is the Localization Quality Focal Loss $L_{LQF}$ (detailed below) from the $LQ$ head; and the third is Weighted Box Regression Loss $L_{WBR}$ (also detailed below) from the $BOX$ head. Importantly, the first two terms together represent the total objectness loss, which can be balanced using the $\lambda_{CLS}$ hyperparameter. By adjusting $\lambda_{CLS} \in [0,1]$, the user can significantly alter the behavior of the resulting model. The smaller $\lambda_{CLS}$ is set, the more the model is incentivized by localization quality, increasing the propensity of detecting diverse OOD objects. During inference-time (and training time to collect proposals for THPN-Box), we use the same linear interpolation to blend the predicted scores from the $CLS$ and $LQ$ heads. The final scores are computed by 
$\mathtt{s = \lambda_{CLS} * cls\_scores + (1 - \lambda_{CLS}) * lq\_scores}$.


\textbf{Localization quality focal loss.~}
A key challenge that proposal networks encounter is data imbalance. The source of imbalance in our case is two-fold: (1) the natural training distribution is often long-tailed, and (2) the pseudo-labels may only cover a handful of samples from each OOD class. By failing to account for this imbalance we risk overfitting the $LQ$ head to the frequently occurring training classes. To combat this, we devise a Localization Quality Focal Loss ($L_{LQF}$) which dynamically weights the loss contribution of each sampled region (i.e., anchor) based on the correctness of the model’s predicted quality score for that region. Specifically, the loss of the $j^{th}$ sampled anchor is:
\begin{equation} \label{eq:lqfl}
L_{LQF}(q_j, q_j^*) = \left |q_j^* - q_j \right |^\gamma L_{BCE}(q_j, q_j^*)
\end{equation}
where $q_j$ and $q_j^*$ are the predicted and target localization quality for the given anchor, respectively. 
$\gamma$ is a hyperparameter to scale the significance of the weighting (we use $\gamma$=2).
While inspired by the original Focal Loss \cite{RetinaNet}, $L_{LQF}$ makes a critical modification to allow it to be used with floating-point targets.
Also, while $L_{LQF}$ bears similarity to the recently proposed QFL \cite{GeneralizedFocalLossV1, GeneralizedFocalLossV2}, the goal of $L_{LQF}$ is different as it encourages accurate localization quality predictions on difficult pseudo-label targets.

\textbf{Weighted box regression loss.~}
Another unique challenge that we face is imperfect pseudo-labels. Particularly when dealing with unseen object categories, it is not safe to assume that the pseudo-label bounding boxes will be of hand-crafted quality. Because box targets are represented as fixed Dirac delta distributions \cite{GeneralizedFocalLossV1} with no encoding of uncertainty, we must be judicious with how much we optimize against certain pseudo-label targets. Naively training on flawed boxes will hinder the model's ability to make fine-grained localization adjustments. 
We address this problem with the $L_{WBR}$ loss, which scales the box regression loss from different pseudo-labels depending on their estimated quality during pseudo-label generation. To scale the loss, we downweight the contribution from anchors matched to pseudo-label targets by the respective pseudo-label's score. Then the loss from the $i^{th}$ anchor is:
\begin{equation} \label{eq:lwbr}
L_{WBR}(t_i, t_i^*) = s_i^\beta L_{1}(t_i, t_i^*).
\end{equation}
Recall, $t_i$ and $t_i^*$ are the predicted and target box offsets, respectively. Here, $s_i \in [0,1]$ is the quality score predicted for pseudo-label $t_i^*$ in Stage 2. Note that for ground-truth targets, we assume $s_i$=1. The hyperparameter $\beta$ scales how severely we downweight the loss from anchors matched to lower scoring targets (we use $\beta$=2). Intuitively, this objective encodes uncertainty into each pseudo-label's box coordinates based on its predicted quality.

\input{tables/coco_cross_cat_generalization.tex}
\input{tables/training_class_diversity_challenge.tex}
\subsection{Implementation details}
\label{sub:implementation_details}
THPN is built on the PyTorch-based \cite{pytorch} mmdetection library \cite{mmdetection}. We use a ResNet-50 \cite{ResNet} with a Feature Pyramid Network (FPN) \cite{FPN} as a backbone. We also use one anchor per feature location and $\lambda_{BOX}=10$ and $\lambda_{BOX}=1$ for THPN-RPN and THPN-Box, respectively, in accordance with Kim et al. \cite{OLN}. Multi-level features from the top scoring anchors are extracted with RoIAlign \cite{MaskRCNN}. In this work, we train all THPN models using crop \& zoom augmentations. We train for $E$=16 epochs initially, and $E$/4=4 epochs in each succeeding self-training round. We use $r$=3 self-training rounds per model and set $p$=30 to incur a 30\% increase in total labels due to pseudo-labels. Note that in each round of self-training, we generate all new pseudo-labels instead of re-using them from previous rounds. Models are trained on four NVIDIA V100 GPUs with a batch-size of two images per device.  

%% file: tables/coco_cross_cat_generalization.tex
\begin{table}[t]

\centering
\resizebox{.7\columnwidth}{!}{

\begin{tabular}{llcc}
\toprule
 &                    & \multicolumn{2}{c}{OOD}       \\
Split   &   Method               & AR10          & AR100         \\ \hline\hline \noalign{\vskip .6mm}
\multirow{9}{*}{VOC} & RPN \cite{FasterRCNN}                 & 7.4           & 20.0          \\
    &   GA-RPN \cite{RegionProposalGuidedAnchoring} & 11.9          & 27.7          \\
    &   Cascade RPN \cite{CascadeRPN}          & 12.6          & 27.7          \\
    &   Faster R-CNN \cite{FasterRCNN}    & 11.6          & 25.1            \\
    &   FCOS \cite{FCOS}                 & 10.5          & 24.4          \\
    &   FCOS-OWP \cite{KevinsPaper}             & 14.5          & 31.3          \\
    &   LDET \cite{LearningToDetectEveryThing}                 & 18.2          & 30.8          \\
    &   OLN \cite{OLN}                 & 18.4          & 33.2          \\ \cdashline{2-4}  \noalign{\vskip .6mm}
    &   \textbf{THPN ($\boldsymbol{\lambda_{CLS}=0}$)}             & \textbf{21.6}          & \textbf{38.9} \\ 

\bottomrule
\end{tabular}

}
\vspace{-2mm}
\caption{Results on the \textit{COCO benchmark} challenge.}
\label{tab:coco_cross_cat_generalization}

\vspace{-4mm}
\end{table}

%% file: tables/training_class_diversity_challenge.tex
\begin{table*}[t]
\centering
\resizebox{0.9\linewidth}{!}{

\begin{tabular}{llc|cccc|cccc|cccc}
\toprule
                        &                                         & Images /    & \multicolumn{4}{c|}{OOD}                                      & \multicolumn{4}{c|}{ID}                                       & \multicolumn{4}{c}{ALL}                                       \\
Split                   & Model                                   & Instances   & AUC           & AR10          & AR100         & AR1k        & AUC           & AR10          & AR100         & AR1k        & AUC           & AR10          & AR100         & AR1k        \\ \hline\hline \noalign{\vskip .6mm}
\multirow{6}{*}{COCO40} & Faster R-CNN                            & 104k / 623k & 26.6          & 17.5          & 36.0          & 51.4          & 44.4          & 41.4          & 58.3          & 63.2          & 39.0          & 33.8          & 51.7          & 60.0          \\
                        & OLN                                     & 104k / 623k & 33.1          & 25.8          & 44.8          & 54.6          & 42.1          & 34.6          & 57.2          & 65.0          & 38.9          & 30.5          & 53.3          & 62.2          \\ \cdashline{2-15} \noalign{\vskip .6mm}
                        & THPN ($\lambda_{CLS}=0$)             & 104k / 810k & 34.1          & 26.9          & 45.9          & \textbf{56.0} & 40.6          & 31.8          & 55.7          & 64.0          & 38.1          & 28.8          & 52.5          & 61.7          \\
                        & THPN ($\lambda_{CLS}=0.10$)          & 104k / 810k & \textbf{34.8} & \textbf{29.8} & \textbf{46.0} & 55.3          & 44.0          & 39.6          & 58.1          & 64.6          & 40.7          & 35.1          & 54.3          & 62.0          \\
                        & \textbf{THPN ($\boldsymbol{\lambda_{CLS}=0.25}$)} & 104k / 810k & 33.6          & 27.6          & 44.4          & 55.1          & 45.6          & 42.8          & 59.5          & 65.2          & \textbf{41.6} & \textbf{37.1} & \textbf{54.9} & 62.4          \\
                        & THPN ($\lambda_{CLS}=0.50$)          & 104k / 810k & 30.9          & 22.5          & 42.1          & 54.5          & \textbf{46.0} & \textbf{43.3} & \textbf{60.2} & \textbf{65.7} & 41.3          & 36.5          & 54.7          & \textbf{62.7} \\ \hline \noalign{\vskip .6mm}
\multirow{6}{*}{VOC}    & Faster R-CNN                            & 95k / 493k  & 19.3          & 11.6          & 25.1          & 42.4          & 46.7          & 45.1          & 60.7          & 64.7          & 34.4          & 29.9          & 44.8          & 55.1          \\
                        & OLN                                     & 95k / 493k  & 24.8          & 18.4          & 33.2          & 45.0          & 44.8          & 40.1          & 59.3          & 66.1          & 35.5          & 29.1          & 47.5          & 56.9          \\ \cdashline{2-15} \noalign{\vskip .6mm}
                        & THPN ($\lambda_{CLS}=0$)             & 95k / 641k  & \textbf{28.8} & 21.6          & \textbf{38.9} & \textbf{49.7} & 43.8          & 37.0          & 58.9          & 65.9          & 36.6          & 28.5          & 49.8          & \textbf{59.0} \\
                        & \textbf{THPN ($\boldsymbol{\lambda_{CLS}=0.10}$)} & 95k / 641k  & 27.9          & \textbf{22.0} & 37.1          & 48.0          & 46.8          & 44.2          & 60.9          & 66.5          & \textbf{38.0} & 32.9          & \textbf{50.2} & 58.5          \\
                        & THPN ($\lambda_{CLS}=0.25$)          & 95k / 641k  & 25.3          & 18.4          & 33.7          & 46.3          & 48.1          & 46.8          & 62.1          & 67.0          & 37.6          & \textbf{33.1} & 49.4          & 58.1          \\
                        & THPN ($\lambda_{CLS}=0.50$)          & 95k / 641k  & 22.2          & 14.7          & 29.6          & 44.8          & \textbf{48.4} & \textbf{47.0} & \textbf{62.8} & \textbf{67.4} & 36.6          & 32.1          & 48.0          & 57.7          \\ \hline \noalign{\vskip .6mm}
\multirow{6}{*}{VOC5}   & Faster R-CNN                            & 74k / 357k  & 16.3          & 9.8           & 20.7          & 38.1          & 48.1          & 47.6          & 62.2          & 65.6          & 29.1          & 24.8          & 37.4          & 49.6          \\
                        & OLN                                     & 74k / 357k  & 20.3          & 14.1          & 26.9          & 40.1          & 47.6          & 45.2          & 61.7          & 67.8          & 31.0          & 25.7          & 40.8          & 51.6          \\ \cdashline{2-15} \noalign{\vskip .6mm}
                        & \textbf{THPN ($\boldsymbol{\lambda_{CLS}=0}$)}    & 74k / 465k  & \textbf{25.6} & \textbf{18.4} & \textbf{34.7} & \textbf{46.6} & 45.8          & 41.1          & 60.6          & 66.9          & \textbf{33.3} & 26.2          & \textbf{44.9} & \textbf{55.1} \\
                        & THPN ($\lambda_{CLS}=0.10$)          & 74k / 465k  & 23.7          & 17.6          & 31.5          & 43.9          & 48.3          & 47.1          & 62.4          & 67.4          & 33.3          & \textbf{28.4} & 43.8          & 53.7          \\
                        & THPN ($\lambda_{CLS}=0.25$)          & 74k / 465k  & 21.8          & 15.5          & 28.6          & 42.3          & 49.4          & 49.2          & 63.5          & 67.7          & 32.7          & 28.3          & 42.6          & 52.9          \\
                        & THPN ($\lambda_{CLS}=0.50$)          & 74k / 465k  & 19.0          & 13.0          & 24.3          & 40.0          & \textbf{49.7} & \textbf{49.5} & \textbf{63.9} & \textbf{68.1} & 31.2          & 27.2          & 40.3          & 51.6          \\ \hline \noalign{\vskip .6mm}
\multirow{6}{*}{Animal} & Faster R-CNN                            & 24k / 63k   & 11.5          & 6.0           & 13.5          & 31.3          & 53.9          & 58.9          & 67.1          & 69.4          & 14.6          & 9.8           & 17.5          & 34.1          \\
                        & OLN                                     & 24k / 63k   & 13.3          & 8.2           & 16.4          & 31.5          & 55.8          & 59.7          & 69.7          & 73.2          & 16.4          & 11.9          & 20.3          & 34.6          \\ \cdashline{2-15} \noalign{\vskip .6mm}
                        & \textbf{THPN ($\boldsymbol{\lambda_{CLS}=0}$)}    & 24k / 81k   & \textbf{18.2} & 10.1          & \textbf{24.9} & \textbf{39.5} & 54.5          & 57.4          & 68.7          & 72.3          & \textbf{20.9} & 13.5          & \textbf{28.1} & \textbf{42.0} \\
                        & THPN ($\lambda_{CLS}=0.10$)          & 24k / 81k   & 17.0          & \textbf{10.3} & 22.9          & 36.6          & 56.1          & 60.6          & 69.9          & 73.0          & 19.8          & \textbf{13.9} & 26.3          & 39.3          \\
                        & THPN ($\lambda_{CLS}=0.25$)          & 24k / 81k   & 16.1          & 10.1          & 20.9          & 35.4          & 56.5          & 61.7          & 70.2          & 73.0          & 19.0          & 13.8          & 24.5          & 38.2          \\
                        & THPN ($\lambda_{CLS}=0.50$)          & 24k / 81k   & 14.7          & 8.7           & 18.7          & 34.3          & \textbf{56.6} & \textbf{61.9} & \textbf{70.3} & \textbf{73.0} & 17.8          & 12.6          & 22.5          & 37.1    
                        \\
\bottomrule
\end{tabular}

}
\vspace{-2mm}
\caption{Results on the \textit{training class diversity} challenge.}
\label{tab:training_class_diversity_challenge}
\vspace{-3mm}
\end{table*}

%% file: sections/experiments.tex
\section{Experiments}
\label{sec:experiments}
To thoroughly evaluate the performance of THPN we consider four generalization challenges which go far beyond the evaluations of contemporary methods \cite{TowardsOpenWorldObjectDetection, KevinsPaper, OLN, LearningToDetectEveryThing}. Our core experimental methodology is to divide the COCO dataset \cite{COCO} into several ID:OOD disjoint class splits, such that the union of the ID and OOD classes equals all 80 COCO classes. During training, we only assume access to labels of the ID classes in the training set.
Importantly, \textit{THPN is only ever exposed to images that contain at least one ID label during training and pseudo-label generation}. Thus, our implementation of THPN does not use any unlabeled training images, just like any non-self-trained baseline. In \cref{sub:coco_benchmark_challenge}, we consider the common VOC$\rightarrow$COCO benchmark.
\cref{sub:training_class_diversity_challenge} and \cref{sub:semi_supervised_challenge} cover our \textit{training class diversity} and \textit{semi-supervised} challenges, respectively. In \cref{sub:ships_challenge}, we test THPN on an open-set ship detection task. Finally, \cref{sub:model_analysis} contains an analysis and ablation study of several model design choices. 

\subsection{COCO benchmark challenge}
\label{sub:coco_benchmark_challenge}
The first challenge we consider is the cross-category generalization task which has been used as the main benchmark by various recent open-world proposal works \cite{OLN, KevinsPaper, LearningToDetectEveryThing}. In this task, we consider the 20 VOC \cite{PASCALVOC} classes to be ID and the 60 remaining (non-VOC) classes to be OOD. We train a model on ID labels only and evaluate by computing Average Recall (AR@$k$ detections per image) \cite{AverageRecall} on the OOD instances in the validation set. We do not consider Average Precision (AP) as it is unfair to penalize false positives unless the dataset is exhaustively labeled. Performance on this task signifies a model's ability to generalize to unseen classes. \cref{tab:coco_cross_cat_generalization} contains the results. We set $\lambda_{CLS}=0$ in this test to maximize OOD performance. THPN outperforms all baselines, surpassing the strongest (OLN) by +3.2\% AR10 and +5.7\% AR100. 
In Appendix B, we evaluate THPN against several learning-free methods such as Selective Search \cite{SelectiveSearch} and EdgeBoxes \cite{EdgeBoxes}, and find that THPN beats the strongest baseline by over 2x. 

\input{tables/semi_supervised_challenge.tex}
\input{tables/ships_challenge.tex}
\subsection{Training class diversity challenge}
\label{sub:training_class_diversity_challenge}
While the \textit{COCO benchmark} challenge provides some notion of a model's open-world aptness, it is a fairly optimistic scenario. Even though there are only 20 training classes, they cover a wide range of COCO's semantic ``superclasses'' like animal, vehicle, and household-object. A model trained on these classes is exposed to a variety of scene types (e.g., indoor, outdoor, etc.), thus improving its generalization \cite{WhatLeadsToGeneralizationOfObjProposals}. Also, while OOD recall alone is important, it does not tell the full story of a model's performance. It is equally critical to measure the model's recall of ID objects, and ultimately the recall of ALL object classes (ID \textit{and} OOD). Our hypothesis is that in the case of strong label bias, existing proposal networks will struggle to generalize to OOD instances without sacrificing ID performance. Meanwhile, THPN's ability to leverage both classification-based and localization-based objectness, as well as high-quality pseudo-labels, will enable it to excel. To test this hypothesis, we curate four ID class splits with increasing difficulty/bias: Half of COCO (COCO40), VOC classes (VOC), a sample of five VOC classes (VOC5), and a highly biased split of only animal classes (Animal). See Appendix H for the exact classes used. Note that AUC serves as summary metric of AR over several $k$ thresholds (10--1000) \cite{OLN}.

\cref{tab:training_class_diversity_challenge} shows the results of this experiment. Note that the results can be interpreted differently depending on the goal of the user. If the goal is to maximize OOD performance, THPN with a small $\lambda_{CLS}$ ($\le 0.25$) outperforms the baselines in all cases. Interestingly, the margins of improvement of OLN over Faster R-CNN \textit{decrease} as we increase label bias (e.g., +8.8\% AR100 on COCO40 down to +2.9\% AR100 on Animal), while THPN's margins over Faster R-CNN \textit{increase} (e.g., +10.0\% AR100 on COCO40 up to +11.4\% AR100 on Animal for $\lambda_{CLS}=0$). This finding confirms our hypothesis that THPN models are far less prone to overfitting than the baselines. If the goal is to maximize ID performance, THPN can also be beneficial. With a larger $\lambda_{CLS}$, THPN can outperform Faster R-CNN on low-bias splits and OLN on high-bias splits. Finally, all THPN variants outperform both baselines in terms of ALL recall, but the choice of $\lambda_{CLS}$ can make a large difference. On COCO40, where 72\% of total instances are from ID classes, users should choose a larger $\lambda_{CLS}$ as ID performance has more influence on ALL recall. On more biased tasks, tasks with more OOD samples than ID samples, or tasks where OOD recall is paramount, a small $\lambda_{CLS}$ is more appropriate. These results showcase the power of allowing the user to influence the ID/OOD tradeoff depending on their needs.

\subsection{Semi-supervised challenge}
\label{sub:semi_supervised_challenge}
Another challenging yet realistic scenario that is not considered by existing open-world detection works is a partially labeled training dataset that only contains labels for a subset of the existing ID instances. In this challenge, we assume a fraction of the original VOC-class instances are labeled. We randomly subsample each class's label count by the same percentage. Our hypothesis is that THPN's self-training  procedure will allow it to generate pseudo-labels on both unlabeled OOD \textit{and} unlabeled ID instances, leading to drastically improved overall recall in these cases.

\cref{tab:semi_supervised_challenge} contains the results of this challenge. We consider having 50\%, 25\%, and 10\% of available labels (to avoid redundancy the 100\% results can be found in \cref{tab:training_class_diversity_challenge} under ``VOC''). In terms of OOD generalization, THPN with $\lambda_{CLS}=0$ performs significantly better than OLN. Importantly, as we reduce the amount of labeled data, THPN's margin of improvement over OLN increases (e.g., +5.7\% AR100 on VOC-100\% up to +6.1\% AR100 on VOC-10\%). For ID performance, we again find a benefit to using a larger $\lambda_{CLS}$ to use a more classification-based objectness. Overall, we find that the best setting to optimize ALL-AUC on all splits is $\lambda_{CLS}=0.10$. Under this setting, a THPN trained with 25\% of labeled samples (and 59\% of images) can outperform a Faster R-CNN trained with all available data! This finding indicates that the optimal $\lambda_{CLS}$ is influenced more by class diversity than label quantity. Overall, we believe our model's ability to gracefully deal with partially labeled datasets is a key advantage.

\begin{figure}[t]
  \centering
   \includegraphics[width=.95\linewidth]{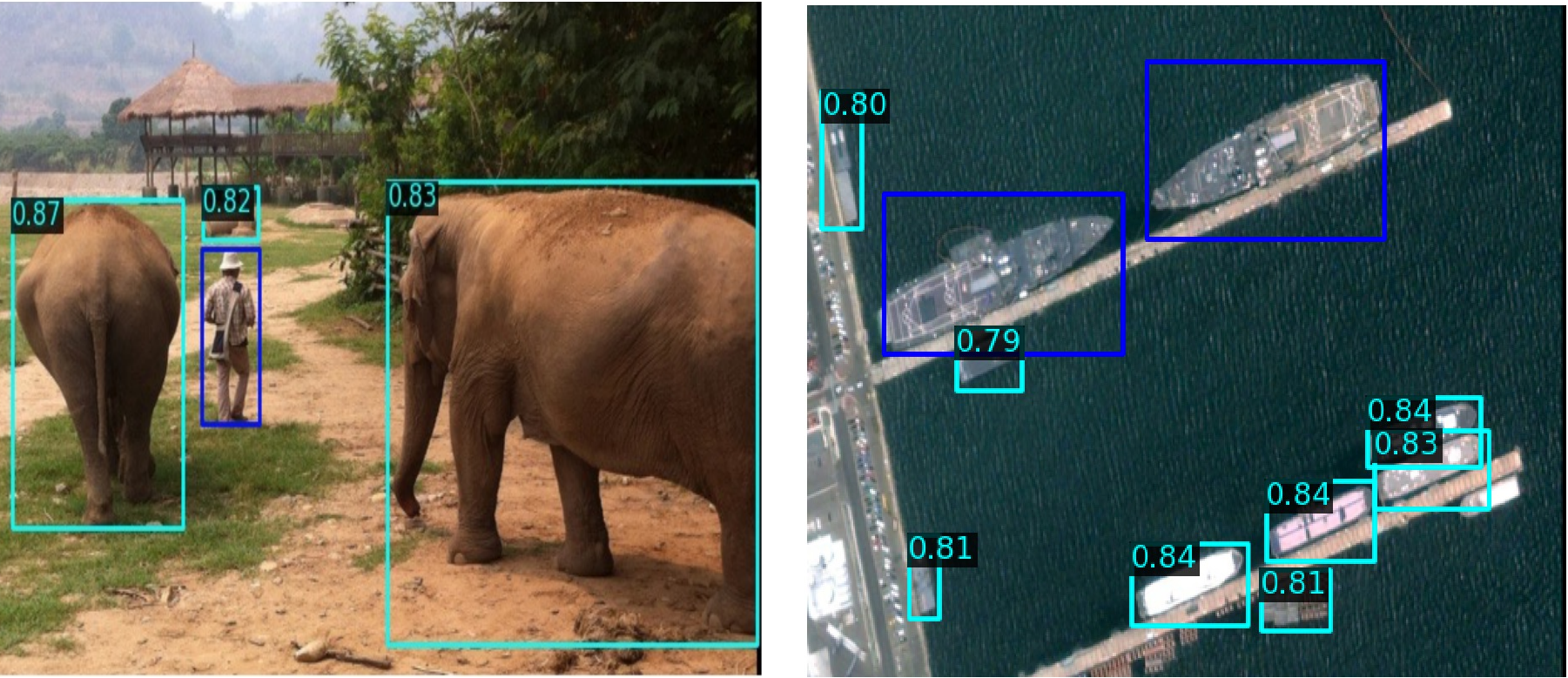}
   \vspace{-2mm}
   \caption{THPN training samples ($p$=30\%). \textcolor{blue}{Blue} boxes are ID labels and \textcolor{cyan}{cyan} boxes are pseudo-labels with objectness $s\in[0,1]$.} 
   \label{fig:pl_samples_main}
   \vspace{-3mm}
\end{figure}
\subsection{Ships challenge}
\label{sub:ships_challenge}
To examine versatility, our final challenge is to consider a domain outside of natural imagery. We use the ShipRSImageNet dataset \cite{ShipRSImageNet}, which contains satellite imagery of oceanic regions around the world, with annotations for both military and civilian/merchant ships. Detection models in this domain are challenged with limited data and significant variations from natural conditions (e.g., reflection, weather, lighting). For this experiment, we consider two splits: (1) train on only military vessels and (2) train on only civilian vessels. From either perspective, being able to detect \textit{all} ships in a given area is a critically important behavior. We argue that this challenge represents a pragmatic view of open-world detection; where the goal is not to detect \textit{every} type of object (e.g., cars, buildings), but only instances of a particular superclass (i.e., ships).

For consistency, we use the same models and experimental settings as our COCO-based challenges. A summary of the results are in \cref{tab:ships_challenge} and the full results are in Appendix E. Note that localization-based objectness is particularly effective for generalizing to OOD ships. Our observation is that in low-data conditions, classification-based objectness learners tend to severely overfit to specific features of the ID classes. For this reason, OLN transcends Faster R-CNN in terms of OOD recall, and $\lambda_{CLS}=0$ is the optimal setting for THPN. This is an eye-opening result as many recent works achieving state-of-the-art performance on ship detection tasks use a variant of Faster R-CNN \cite{ShipRSImageNet, lightweight_fasterrcnn_for_ship_detection, ship_detection_fasterrcnn_sar, rcnn_based_ship_detection}. While localization-based objectness provides one advantage, our self-training procedure boosts performance much higher than a vanilla OLN. On the Civilian split, THPN improves OOD recall over OLN and Faster R-CNN by +11.1 and +33.8 AUC, respectively! In terms of ID recall, THPN with a small $\lambda_{CLS}$ can slightly exceed OLN. As a result, THPN's recall on ALL ships is vastly superior to either baseline. This result portrays THPN's capability across domains, and call into question the common algorithms used for remote sensing detection.

\subsection{Analysis of model design}
\label{sub:model_analysis}

\begin{figure}[t]
  \centering
   \includegraphics[width=0.9\linewidth]{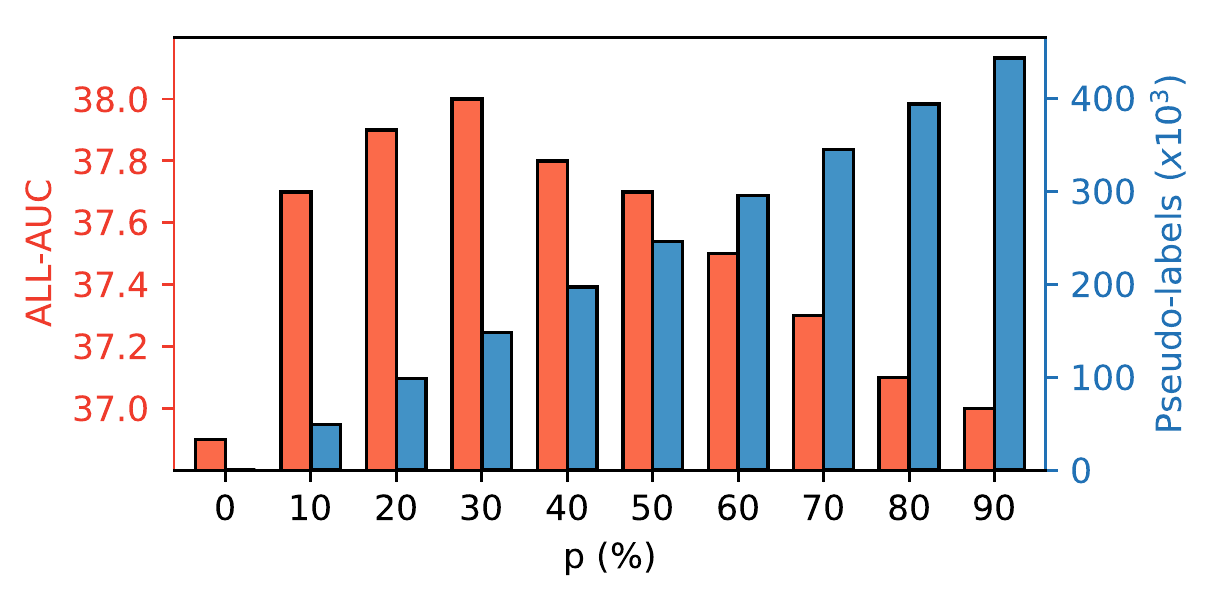}
   \vspace{-3mm}
   \caption{Hyperparameter $p$'s impact on ALL recall and pseudo-label count for a THPN trained on the VOC split.}
   \label{fig:hyperparameter_p}
   \vspace{-3mm}
\end{figure}

\input{tables/ablation.tex}

Training a THPN involves making a few key design decisions. First, we explore the implications of hyperparameter $p$ (i.e., how many pseudo-labels to allow) on ALL recall for the VOC split in \cref{fig:hyperparameter_p}. We find that using $p=30\%$ is a good rule of thumb in all scenarios, though the performance is not very sensitive to $p$. We visualize two pseudo-labeled samples from THPN with $p$=30\% in \cref{fig:pl_samples_main}. Notice that the pseudo-labels are of high quality in both domains, achieving a reasonable balance of recall and precision on OOD objects. See Appendix F for more pseudo-labeled samples.

\cref{tab:ablation} contains an ablation study for THPN on the VOC split. Self-training is responsible primarily for improved OOD performance, however THPN ($\lambda_{CLS} = 0.10$) without self-training still outperforms OLN significantly (OLN achieves 24.8 OOD-AUC, 44.8 ID-AUC, 35.5 ALL-AUC). As expected, removing the $LQ$ head ($\lambda_{CLS}=1$) results in much worse OOD recall with a slight benefit to ID recall, while removing the $CLS$ head ($\lambda_{CLS}=0$) yields worse ID recall but improved OOD performance. Our two dynamic loss functions, $L_{LQF}$ and $L_{WBR}$, also play an important role especially for OOD detection. In additional experiments we find that these losses become more important in more biased class splits. For example, on the Animal split, removing these two losses leads to a 2.4 OOD-AUC reduction. Interestingly, we find that our finetuning-based self-training is not only more efficient, but better performing than the conventional retraining-based approach. Finally, while the longer (initial) 16-epoch training schedule is beneficial for both OOD and ID recall, the crop \& zoom augmentations mainly benefit OOD generalization. 

%% file: tables/semi_supervised_challenge.tex
\begin{table*}[t]
\centering
\resizebox{0.9\linewidth}{!}{

\begin{tabular}{llc|cccc|cccc|cccc}
\toprule
                            &                                         & Images /   & \multicolumn{4}{c|}{OOD}                                      & \multicolumn{4}{c|}{ID}                                       & \multicolumn{4}{c}{ALL}                                       \\
Split                       & Model                                   & Instances  & AUC           & AR10          & AR100         & AR1k          & AUC           & AR10          & AR100         & AR1k          & AUC           & AR10          & AR100         & AR1k          \\ \hline\hline \noalign{\vskip .6mm}
\multirow{6}{*}{VOC (50\%)} & Faster R-CNN                            & 75k / 246k & 18.7          & 11.7          & 24.1          & 40.9          & 44.8          & 42.7          & 58.5          & 63.1          & 33.1          & 28.5          & 43.2          & 53.6          \\
                            & OLN                                     & 75k / 246k & 23.8          & 17.7          & 31.7          & 43.8          & 44.4          & 39.5          & 58.8          & 65.7          & 34.9          & 28.5          & 46.7          & 56.3          \\ \cdashline{2-15} \noalign{\vskip .6mm}
                            & THPN ($\lambda_{CLS}=0$)             & 75k / 320k & \textbf{27.9} & \textbf{21.2} & \textbf{37.4} & \textbf{48.2} & 43.7          & 38.6          & 58.0          & 65.0          & 36.2          & 29.4          & \textbf{48.7} & \textbf{57.8} \\
                            & \textbf{THPN ($\boldsymbol{\lambda_{CLS}=0.10}$)} & 75k / 320k & 25.7          & 19.6          & 34.0          & 45.9          & 46.1          & 44.1          & 59.7          & 65.7          & \textbf{36.7} & \textbf{32.0} & 48.1          & 57.2          \\
                            & THPN ($\lambda_{CLS}=0.25$)          & 75k / 320k & 23.8          & 16.8          & 31.6          & 45.2          & 47.1          & \textbf{45.7} & 60.8          & 66.1          & 36.5          & 32.0          & 47.7          & 57.1          \\
                            & THPN ($\lambda_{CLS}=0.50$)          & 75k / 320k & 21.3          & 13.8          & 28.1          & 43.7          & \textbf{47.1} & 45.4          & \textbf{61.1} & \textbf{66.3} & 35.5          & 30.9          & 46.4          & 56.6          \\ \hline \noalign{\vskip .6mm}
\multirow{6}{*}{VOC (25\%)} & Faster R-CNN                            & 56k / 123k & 17.9          & 11.2          & 22.9          & 39.2          & 42.7          & 40.1          & 55.8          & 60.9          & 31.6          & 27.0          & 41.1          & 51.6          \\
                            & OLN                                     & 56k / 123k & 21.9          & 16.6          & 28.8          & 40.7          & 43.2          & 38.3          & 57.1          & 64.1          & 33.4          & 27.5          & 44.5          & 54.0          \\ \cdashline{2-15} \noalign{\vskip .6mm}
                            & THPN ($\lambda_{CLS}=0$)             & 56k / 160k & \textbf{26.2} & \textbf{19.7} & \textbf{34.9} & \textbf{46.3} & 43.1          & 38.6          & 56.9          & 64.0          & 35.2          & 28.9          & \textbf{47.0} & \textbf{56.4} \\
                            & \textbf{THPN ($\boldsymbol{\lambda_{CLS}=0.10}$)} & 56k / 160k & 24.2          & 17.9          & 32.1          & 44.8          & 44.7          & 42.4          & 58.0          & 64.2          & \textbf{35.3} & 30.5          & 46.4          & 55.9          \\
                            & THPN ($\lambda_{CLS}=0.25$)          & 56k / 160k & 22.7          & 15.9          & 30.2          & 43.7          & 45.5          & \textbf{43.9} & 58.8          & 64.4          & 35.1          & \textbf{30.6} & 46.0          & 55.5          \\
                            & THPN ($\lambda_{CLS}=0.50$)          & 56k / 160k & 20.6          & 13.4          & 26.9          & 43.0          & \textbf{45.6} & 43.6          & \textbf{59.0} & \textbf{64.7} & 34.3          & 29.7          & 44.7          & 55.4          \\ \hline \noalign{\vskip .6mm}
\multirow{6}{*}{VOC (10\%)} & Faster R-CNN                            & 33k / 49k  & 16.2          & 10.4          & 20.5          & 35.8          & 39.5          & 36.2          & 51.8          & 57.8          & 29.1          & 24.5          & 37.9          & 48.4          \\
                            & OLN                                     & 33k / 49k  & 19.8          & 15.2          & 25.7          & 37.3          & 40.8          & 36.3          & 53.6          & 61.0          & 31.3          & 26.0          & 41.3          & 50.8          \\ \cdashline{2-15} \noalign{\vskip .6mm}
                            & THPN ($\lambda_{CLS}=0$)             & 33k / 64k  & \textbf{24.0} & \textbf{18.1} & \textbf{31.8} & \textbf{43.5} & 41.0          & 36.5          & 54.0          & 61.6          & 33.1          & 27.1          & \textbf{44.0} & \textbf{53.9} \\
                            & \textbf{THPN ($\boldsymbol{\lambda_{CLS}=0.10}$)} & 33k / 64k  & 23.0          & 17.2          & 30.2          & 42.4          & 42.3          & 39.5          & 55.0          & 61.9          & \textbf{33.5} & \textbf{28.6} & 43.9          & 53.6          \\
                            & THPN ($\lambda_{CLS}=0.25$)          & 33k / 64k  & 20.9          & 14.8          & 27.4          & 40.5          & 43.0          & \textbf{40.9} & 55.5          & 62.0          & 33.0          & 28.6          & 43.0          & 52.8          \\
                            & THPN ($\lambda_{CLS}=0.50$)          & 33k / 64k  & 19.0          & 12.4          & 25.0          & 39.7          & \textbf{43.0} & 40.5          & \textbf{55.8} & \textbf{62.1} & 32.3          & 27.6          & 42.1          & 52.5       
                        \\
\bottomrule
\end{tabular}

}
\vspace{-2mm}
\caption{Results on the \textit{semi-supervised} challenge.}
\label{tab:semi_supervised_challenge}
\vspace{-2mm}
\end{table*}

%% file: tables/ships_challenge.tex
\begin{table}[t]
\centering
\resizebox{.9\columnwidth}{!}{

\begin{tabular}{llcccc}
\toprule
                          &                                      & Images /    & OOD           & ID            & ALL           \\
Split                     & Model                                & Instances   & AUC           & AUC           & AUC           \\ \hline\hline \noalign{\vskip .6mm}
\multirow{4}{*}{Military} & Faster R-CNN                         & 1.5k / 4.7k & 11.8          & 65.6          & 33.3          \\
                          & OLN                                  & 1.5k / 4.7k & 23.0          & 69.0          & 41.3          \\ \cdashline{2-6} \noalign{\vskip .9mm}
                          & \textbf{THPN ($\boldsymbol{\lambda_{CLS}=0}$)} & 1.5k / 6.1k & \textbf{29.0} & 68.8          & \textbf{44.7} \\
                          & THPN ($\lambda_{CLS}=0.10$)       & 1.5k / 6.1k & 26.6          & \textbf{69.3} & 43.6          \\ \hline \noalign{\vskip .6mm}
\multirow{4}{*}{Civilian} & Faster R-CNN                         & 1.0k / 4.4k & 16.0          & 33.4          & 24.6          \\
                          & OLN                                  & 1.0k / 4.4k & 38.7          & 35.6          & 36.9          \\ \cdashline{2-6} \noalign{\vskip .9mm}
                          & \textbf{THPN ($\boldsymbol{\lambda_{CLS}=0}$)} & 1.0k / 5.7k & \textbf{49.8} & \textbf{36.5} & \textbf{42.8} \\
                          & THPN ($\lambda_{CLS}=0.10$)       & 1.0k / 5.7k & 46.3          & 36.4          & 41.0      
\\
\bottomrule
\end{tabular}

}
\vspace{-2mm}
\caption{Results on the \textit{ships} challenge.}
\label{tab:ships_challenge}
\vspace{-3mm}
\end{table}

%% file: tables/ablation.tex
\begin{table}[t]
\centering
\resizebox{.9\columnwidth}{!}{

\begin{tabular}{lccc}
\toprule
Model                                & OOD-AUC      & ID-AUC      & ALL-AUC     \\ \hline\hline \noalign{\vskip .6mm}
THPN ($\lambda_{CLS}=0.10$)                                  & 27.9         & 46.8        & 38.0        \\
No self-training                      & 25.8 \textcolor{red}{(-2.1)}    & 46.5 \textcolor{red}{(-0.3)}    & 36.9 \textcolor{red}{(-1.1)} \\
No \textit{LQ} head                   & 20.3 \textcolor{red}{(-7.6)}    & 47.9 \textcolor{blue}{(+1.1)}    & 35.6 \textcolor{red}{(-2.4)} \\
No \textit{CLS} head                  & 28.8 \textcolor{blue}{(+0.9)}  & 43.8 \textcolor{red}{(-3.0)}      & 36.6 \textcolor{red}{(-1.4)} \\
No $L_{LQF}$                          & 27.2 \textcolor{red}{(-0.7)}    & 46.1 \textcolor{red}{(-0.7)}      & 37.3 \textcolor{red}{(-0.7)} \\
No $L_{WBR}$                          & 27.5 \textcolor{red}{(-0.4)}    & 46.6 \textcolor{red}{(-0.2)}      & 37.6 \textcolor{red}{(-0.4)} \\
Finetune $\rightarrow$ Retrain                   & 27.4 \textcolor{red}{(-0.5)}    & 46.0 \textcolor{red}{(-0.8)}      & 37.3 \textcolor{red}{(-0.7)} \\
1x schedule                           & 27.4 \textcolor{red}{(-0.5)}    & 46.1 \textcolor{red}{(-0.7)}      & 37.4 \textcolor{red}{(-0.6)} \\
No data aug.                          & 25.1 \textcolor{red}{(-2.8)}    & 47.5 \textcolor{blue}{(+0.7)}    & 37.2 \textcolor{red}{(-0.8)} \\

\bottomrule
\end{tabular}

}
\vspace{-2mm}
\caption{Ablation study on the VOC split.}
\label{tab:ablation}

\vspace{-3mm}
\end{table}

%% file: sections/conclusion.tex
\section{Conclusion}

In the scope of open-world detection tasks, the variation of data bias and desired model behavior renders static proposal networks insufficient. In this work, we instead introduce a powerful new class of proposal solution that can be easily adjusted to suit the gamut of challenging open-world scenarios. Our novel evaluation challenges test models in a variety of conditions, ranging from large-scale academic tasks, to tasks with severe degrees of ID class bias and partial labels. We also demonstrate our model's superiority in realistic remote sensing applications. THPN's superior recall of both ID and OOD objects has the potential to enhance a variety of open-world applications, and we hope that our evaluation protocols can serve as touchstones to inspire even more robust models in the future. 

\section*{Acknowledgements}
This work is supported by NSF CNS-2112562, IIS-2140247, and AFRL FA8750-21-1-1015.

%% file: sections/appendix.tex
\clearpage
\onecolumn
\section*{Appendix}

\begin{figure*}[t]
  \centering
   \includegraphics[width=0.9\linewidth]{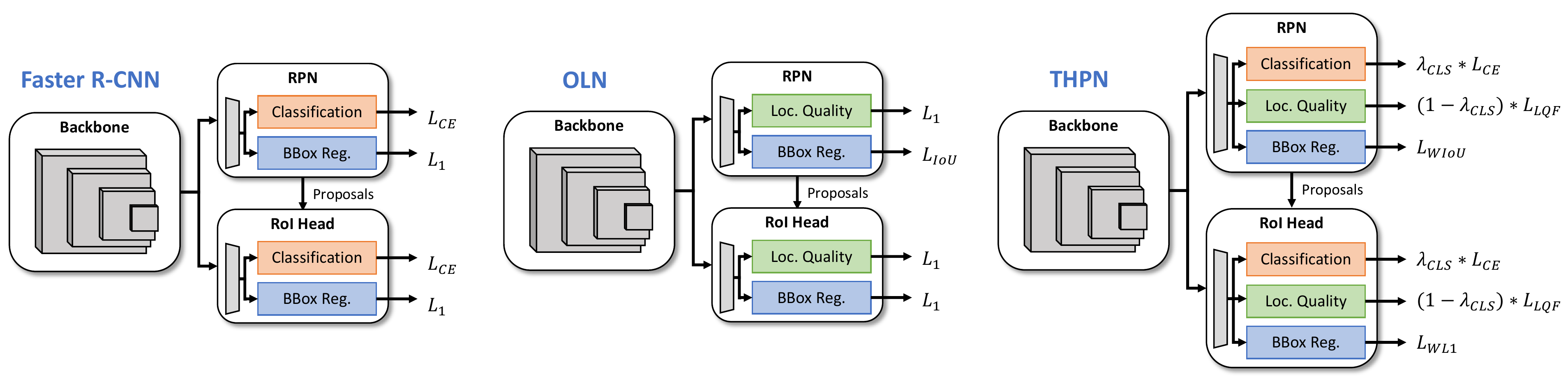}
   \vspace{-2mm}
   \caption{Differences between Faster R-CNN, OLN, and our hybrid THPN architecture.}
   \label{fig:architecture_diffs}
\end{figure*}

\subsection*{A. Overview of baseline methods}
There are currently three types of object proposals methods: (1) learning-free approaches, (2) classification-based learning approaches, and (3) pure localization-based learning approaches. In this section, we outline the methodology for each and discuss their respective drawbacks. See \cref{fig:architecture_diffs} for a visual representation of the architectural differences between our approach and existing proposal networks.
\label{appendix:baselines}

\paragraph{Learning-free approaches.}
Early approaches for generating object proposals rely on hand-crafted features such as contrast and edges. Krähenbühl and Koltun \cite{GeodesicObjectProposals} identify critical level sets in geodesic distance transforms computed over carefully placed seeds to generate proposals.  Uijlings et al.\cite{SelectiveSearch} propose Selective Search, an algorithm that performs a graph-based segmentation and greedily merges superpixels based on color, texture, size, and shape similarity. Edge Boxes \cite{EdgeBoxes} is a box proposal algorithm that works by efficiently computing and finding enclosed edge contours in an image. Finally, Multiscale Combinatorial Grouping (MCG) \cite{MCG} performs multi-scale segmentation and groups regions by efficiently exploring their combinatorial space. These methods were once the backbone of many early object detection systems such as Fast R-CNN \cite{FastRCNN}, but have since been outclassed by learning-based approaches in terms of efficiency and recall performance.

\paragraph{Classification-based learning approaches.}
The seminal learning-based proposal solution is the Region Proposal Network (RPN) \cite{FasterRCNN}. For a predefined set of anchor boxes, RPN learns to predict box coordinate regression deltas as well as an ``objectness" score which indicates the likelihood that the box contains a foreground object of interest as opposed to background or an object from a class outside of the training distribution.  While the single-stage RPN serves as an efficient benchmark solution for many state-of-the-art detection systems \cite{FasterRCNN, CascadeRCNN, MaskRCNN}, we primarily focus on two-stage approaches in this work as they have been shown to have significantly better generalization abilities \cite{WhatLeadsToGeneralizationOfObjProposals, TowardsOpenWorldObjectDetection, OLN}. Thus, we consider a class-agnostic Faster R-CNN model as a stronger baseline, in which all annotations are treated as instances of the same class. The loss function for both stages of a class agnostic Faster R-CNN model (i.e., RPN and RoI Head) for an image is defined as:
\begin{equation} \label{eq:rpn_loss}
\begin{split}
    L_{RPN}\big(\{c_i\}, \{t_i\}\big) = \frac{1}{N_{cls}} \sum_i L_{BCE}(c_i, c_i^*) + \lambda \frac{1}{N_{reg}} \sum_i c_i^* L_{1}(t_i, t_i^*).
\end{split}
\end{equation}
Here, $i$ is the anchor index, $c_i$ is the predicted probability of anchor $i$ containing an object, and $t_i$ are the predicted box regression deltas for anchor $i$. The ground truth labels for objectness and box regression are shown as $c_i^*$ and $t_i^*$, respectively. The box regression head is trained with an $L_1$ loss. The objectness head of RPN is trained with a binary cross-entropy loss $L_{BCE}$, and a sampler is used to ensure 50\% of boxes sampled during training are positive matches to some ground truth object.

Kim et al. \cite{OLN} point out that framing object proposal as a discriminative task hinders generalization as it involves indiscriminate sampling of negative regions when training the objectness head(s). In other words, because we only have access to labels from a subset of object classes that exist in the world, the model overfits to the labeled object classes and treats all unlabeled objects as background.

\paragraph{Pure localization-based learning approach.}
Object Localization Network (OLN) \cite{OLN} provides an alternative to classification-based approaches in an effort to resolve the aforementioned overfitting issue. OLN uses the same architecture as Faster R-CNN, but it replaces the classification heads with localization quality prediction heads. OLN uses centerness \cite{FCOS} and IoU score \cite{IoUNet} as substitutes for objectness in the OLN-RPN and OLN-Box stages, respectively. Thus, OLN is a pure localization-based proposal network trained with the following loss function:
\begin{equation} \label{eq:oln_loss}
\begin{split}
    L_{OLN}\big(\{q_j\}, \{t_i\}\big) = \frac{1}{N_{lq}} \sum_j L_1(q_j, q_j^*) + \lambda \frac{1}{N_{reg}} \sum_i L_1(t_i, t_i^*).
\end{split}
\end{equation}
Here, $q_j$ and $q_j^*$ are the predicted and target localization quality scores, respectively. Note that we use a separate index $j$ because the set of sampled anchors for the $LQ$ head can be different than the set of sampled anchors for the $BOX$ head. Because we are not training a discriminative classification task, we only need to sample positively matched anchors to train the $LQ$ head. Intuitively, if a model can accurately predict its overlap $q_j \in [0,1]$ with a ground truth object, we can effectively treat $q_j$ as a notion of objectness. This re-framing of the sparse classification problem allows the model to be less biased towards the specific classes that it is trained on.

While OLN resolves the explicit bias resulting from learning to classify all unlabeled regions as background, we posit that it still suffers from implicit bias because it still only learns from ID instances. Adding to the bias problem is the fact that there exists a natural class imbalance issue in natural data. While OLN's generalization benefits over Faster R-CNN is shown to be significant when it is trained on a diverse and representative class set such as PASCAL VOC \cite{PASCALVOC}, we hypothesize that it will struggle when faced with more challenging tasks with fewer ID classes and fewer labeled instances.

\subsection*{B. Learning-free baseline comparison}
\input{tables/learning_free_baselines.tex}
While learning-free approaches to object proposal have been largely supplanted by deep learning-based methods, we feel that they are still worth worth comparing against. For this test, we compare recall on all classes in the COCO validation set. For this experiment, we use a THPN with $\lambda_{CLS} = 0.50$ to promote the optimal ID recall. \cref{tab:learning_free_baselines} shows that THPN outclasses all learning-free baselines by a significant margin.

\subsection*{C. Closed-set performance}
\input{tables/closed_set.tex}
THPN's design is primarily suited for handling open-set tasks that assume the presence of OOD objects of interest. However, we notice that a THPN with a larger $\lambda_{CLS}$ is capable of superior ID performance to pure classification-based models like Faster R-CNN. For completeness, we test our model's performance in closed-set tasks. The results of this test are in \cref{tab:closed_set}. We train each model on all classes of COCO or ShipRSImageNet (except \textit{docks}) and test on the complete validation sets. Note that we do not use self-training or crop \& zoom augmentations here to test the impact of the architectural differences only. On the large-scale COCO dataset, Faster R-CNN outperforms OLN. However, THPN with $\lambda_{CLS} \ge 0.10$ beats both baselines. Setting $\lambda_{CLS} = 0.50$ is the best in this case, beating Faster R-CNN by +2.4 AUC. On the smaller ships data, OLN is superior to Faster R-CNN, and THPN with $\lambda_{CLS} \le 0.25$ outperforms OLN by +0.8 AUC.

\subsection*{D. Data augmentation and training schedule}
\input{tables/augmentations.tex}
\cref{tab:augmentations} shows the effect of various data augmentations and training schedules on an OLN model trained on the VOC split. In this work, we borrow transform implementations from mmdetection \cite{mmdetection}. We find that the crop \& zoom augmentation is the most effective for improving OOD generalization of an OLN-style of proposal network. However, all augmentations come at a slight cost of ID recall. We also find that using a 2x training schedule (16 epochs) is beneficial only if using the additional strong augmentations. 

\subsection*{E. Full ship detection results}
\input{tables/ships_challenge_full.tex}
For the purposes of space efficiency, we show a summary of the results from the \textit{ships} challenge in \cref{sub:ships_challenge} of the main text. In this section, we provide the full results of this experiment in \cref{tab:ships_challenge_full}. In both splits, THPN with $\lambda_{CLS}=0$ is vastly superior to the baselines in terms of OOD recall (regardless of $k$ in AR@$k$). Because OLN also effectively has $\lambda_{CLS}=0$, it is clear that our self-training procedure is very effective in this domain. This is likely due to the fact that there are relatively few labeled training instances compared to COCO. Given the correct setting of $\lambda_{CLS}$, THPN can also slightly improve the ID recall on both splits. 

\subsection*{F. Inference-time dynamic $\lambda_{CLS}$}
\input{tables/tl10_training_class_diversity_challenge.tex}
\input{tables/tl10_semi_supervised_challenge.tex}
In all other experiments, the $\lambda_{CLS}$ associated with the THPN models is consistent during both training (i.e., loss weighting and score weighting during pseudo-label generation) and inference (score weighting). For the sake of completeness, we include results where we use one $\lambda_{CLS}$ for training, and alter it during inference. This implementation of THPN makes the model truly inference-time dynamic. We evaluate on the \textit{training class diversity} and \textit{semi-supervised} challenges using $\lambda_{CLS}=0.10$ during training. The results of these experiments are in Table \ref{tab:tl10_training_class_diversity_challenge} and Table \ref{tab:tl10_semi_supervised_challenge}, respectively. Overall, we find that this inference-time-dynamic variant is capable of approaching the performance of the static-trained variant in terms of ALL recall. However, it achieves this by creating a more optimal balance between OOD recall and ID recall, rather than improving either one individually. The dynamic variants have much less deviation in performance between $\lambda_{CLS}$ values. Interestingly, the most optimal setting (in terms of ALL recall) for a dynamic THPN using $\lambda_{CLS}=0.10$ during training is to use $\lambda_{CLS}=0.25$ during inference. Regardless, these experiments confirm that THPN can be treated as dynamic at inference-time to good effect.

\subsection*{G. Pseudo-label visualizations}
\begin{figure*}[h]
  \centering
   \includegraphics[width=0.85\linewidth]{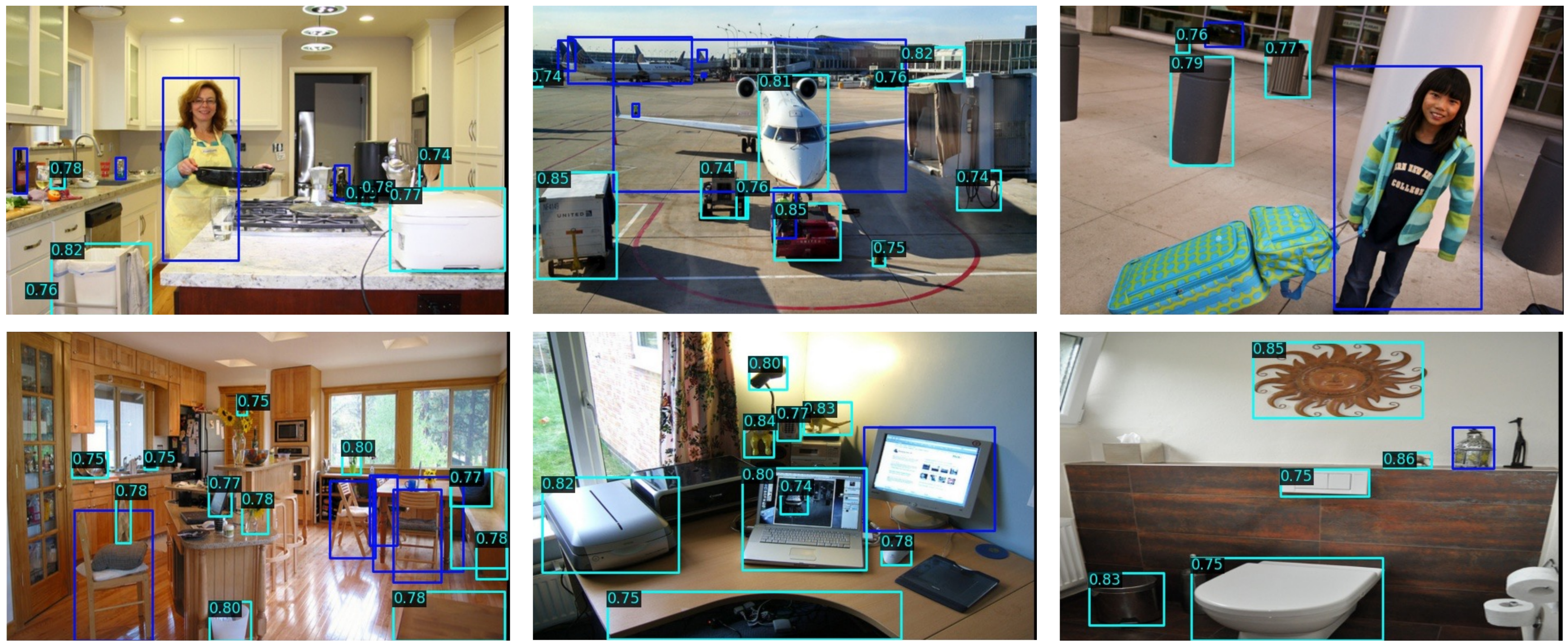}
   \caption{Visualization of pseudo-label boxes \textcolor{cyan}{(cyan)} alongside the original ground truth boxes \textcolor{blue}{(blue)} on various training images. Here, we consider the VOC split and the pseudo-labels are from a THPN ($\lambda_{CLS} = 0.10$) model with $p$=30\%.}
   \label{fig:pl_samples_voc}
\end{figure*}

\begin{figure*}[h]
  \centering
   \includegraphics[width=0.85\linewidth]{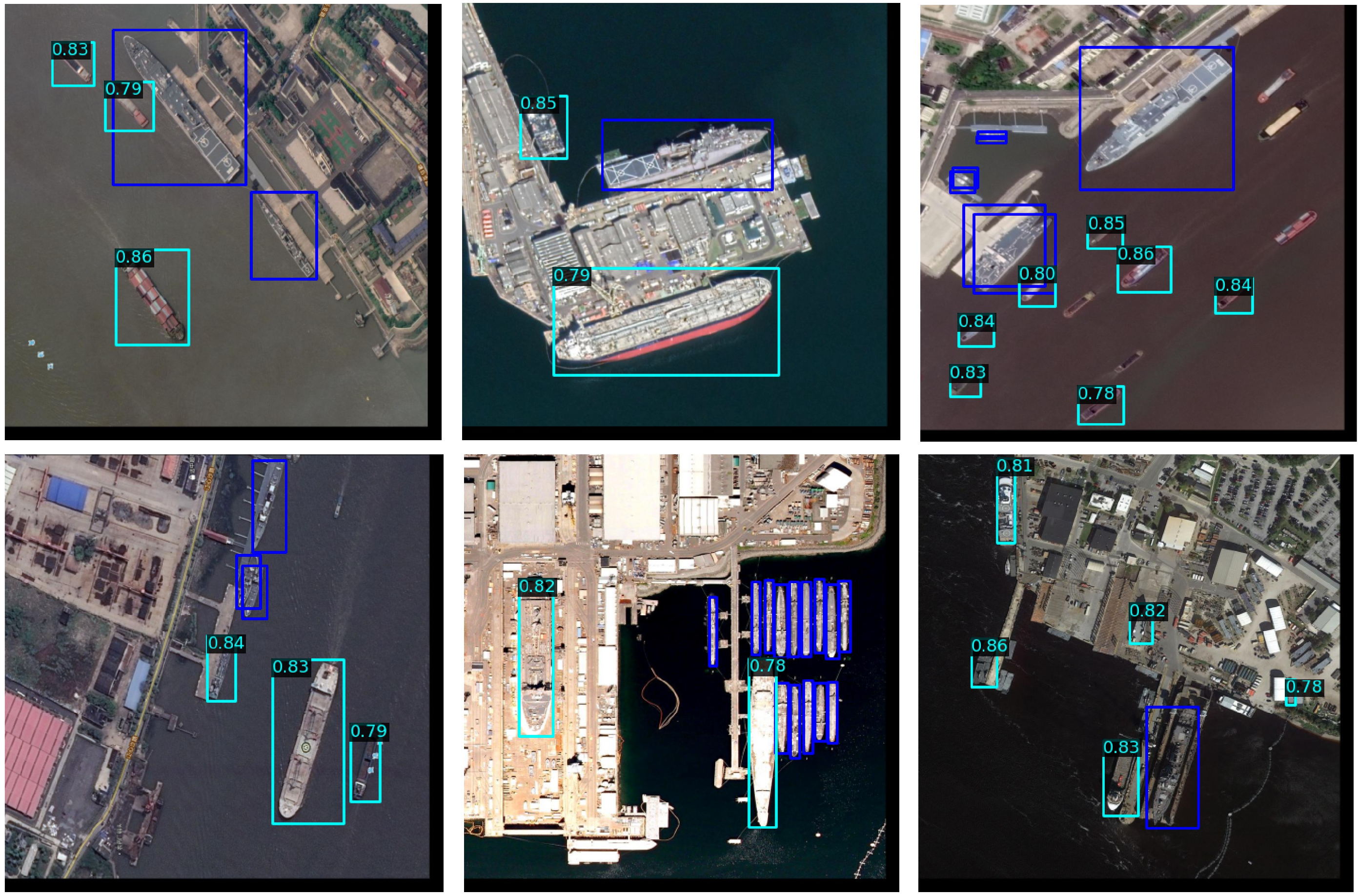}
   \caption{Visualization of pseudo-label boxes \textcolor{cyan}{(cyan)} alongside the original ground truth boxes \textcolor{blue}{(blue)} on various training images. Here, we consider the Warship split and the pseudo-labels are from a THPN ($\lambda_{CLS} = 0$) model with $p$=30\%.}
   \label{fig:pl_samples_warship}
\end{figure*}

\subsection*{H. Prediction visualizations}
\begin{figure*}[h]
  \centering
   \includegraphics[width=0.85\linewidth]{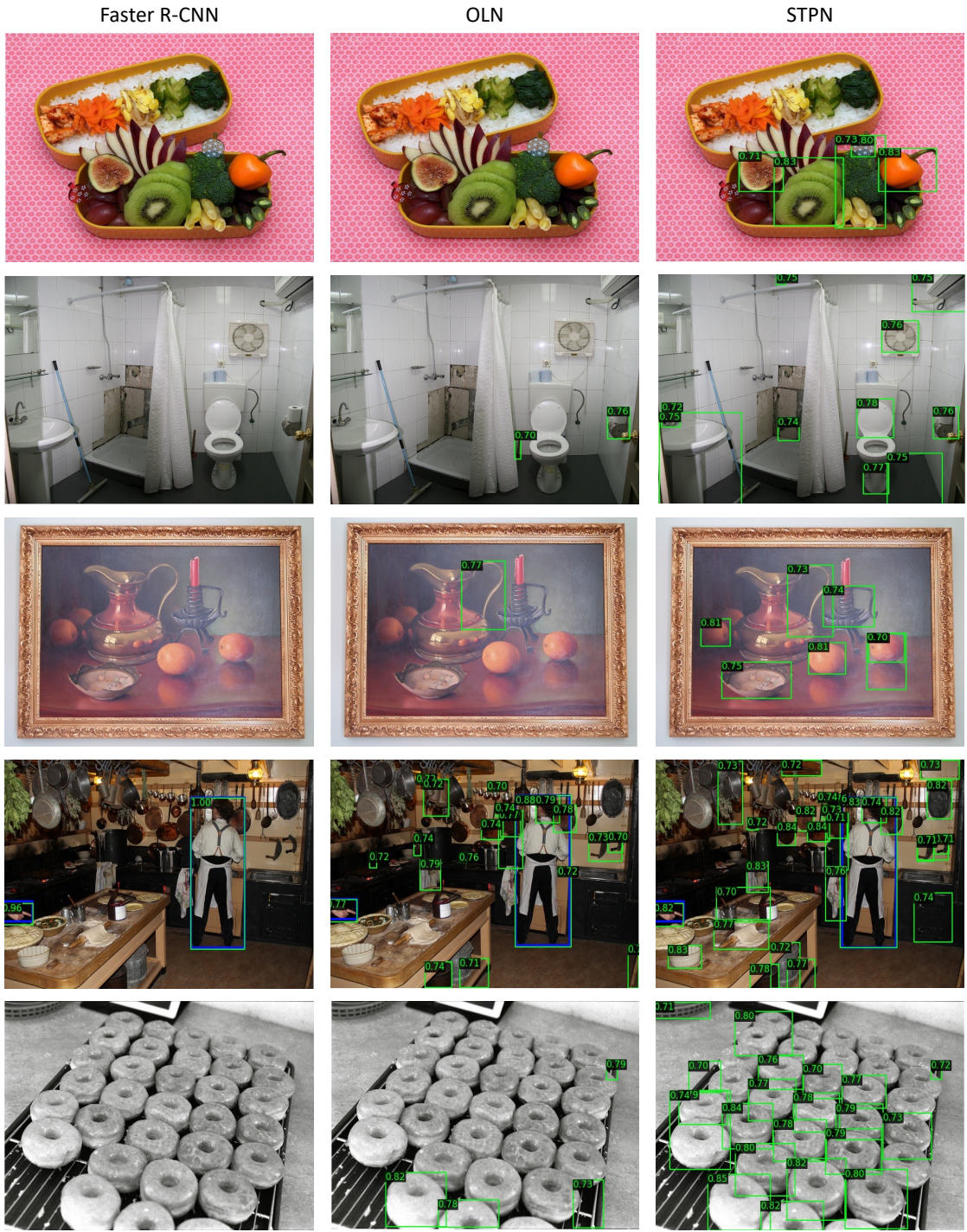}
   \caption{Visualization of predicted proposals \textcolor{green}{(green)} alongside the ground truth boxes \textcolor{blue}{(blue)} on various COCO validation images. All models are trained on the VOC5 split (THPN is trained with $\lambda_{CLS} = 0$). In general, THPN is able to detect many OOD objects that Faster R-CNN and OLN misses. We find that a common failure mode of localization-based models is proposing parts of a larger object (e.g., a person's hand, the leg of a chair, etc.).}
   \label{fig:voc_predictions}
\end{figure*}

\clearpage
\begin{figure*}[h]
  \centering
   \includegraphics[width=0.85\linewidth]{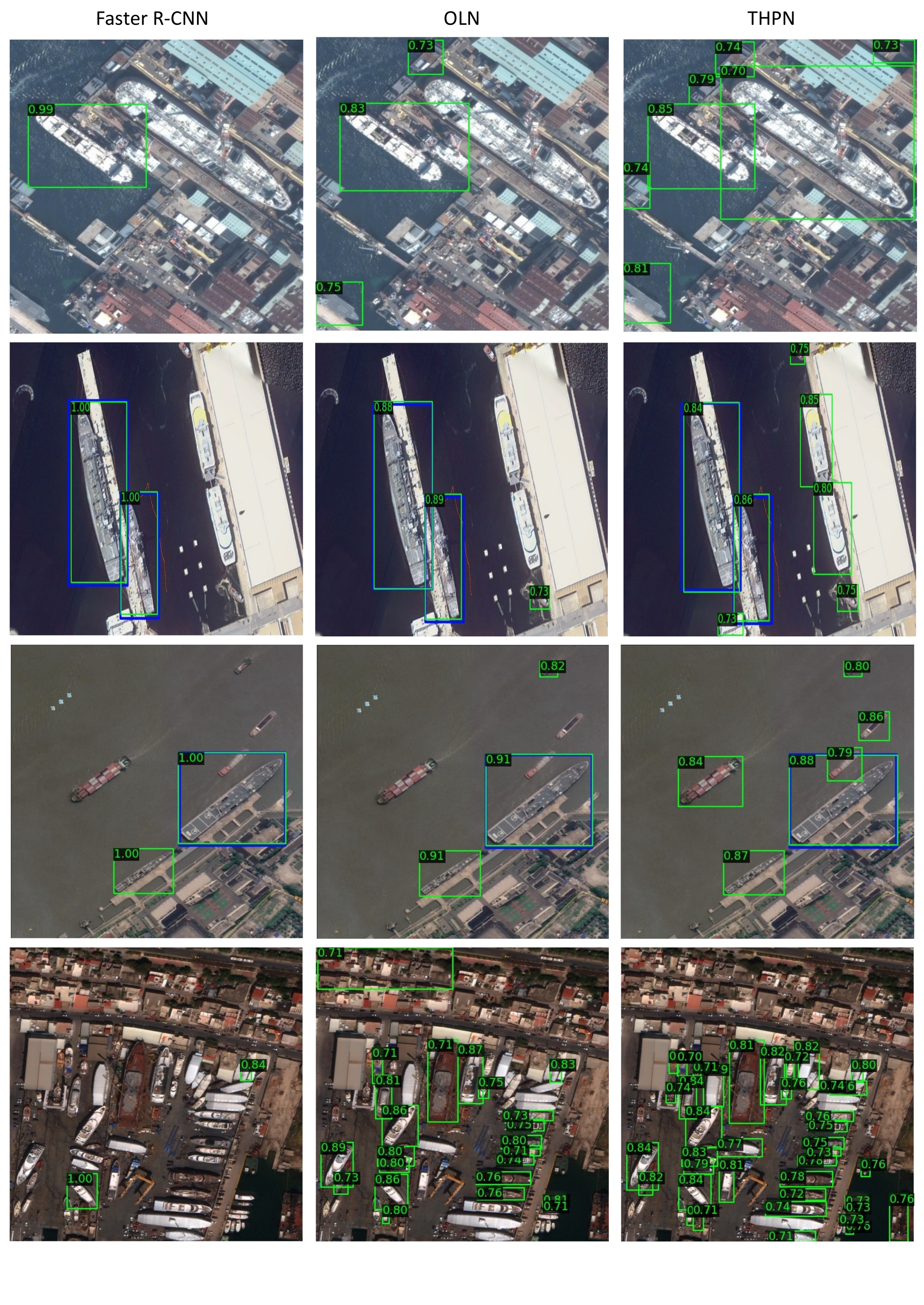}
   \caption{Visualization of predicted proposals \textcolor{green}{(green)} alongside the ground truth boxes \textcolor{blue}{(blue)} on various ShipRSImageNet validation images. All models are trained on the Military split (THPN is trained with $\lambda_{CLS} = 0$). In general, THPN is able to detect many more OOD ships than Faster R-CNN and OLN while maintaining performance on the ID military ships. We find that a common failure mode of THPN on this dataset is to localize large defined objects on shore (e.g., storage containers, docks, etc.).}
   \label{fig:warship_predictions}
\end{figure*}

\clearpage
\subsection*{I. Training class diversity challenge splits}
\input{tables/coco_splits.tex}

\clearpage
\subsection*{J. Frequently Asked Questions (FAQ)}
\begin{itemize}
  \item Why do we call this method ``for the Open World"?
  \begin{itemize}
      \item According to the terminology in the literature, ``open-set recognition" refers to the task of dealing with OOD inputs \cite{OGOpenSet}, while ``open-world recognition" refers to the task of detecting OOD inputs and incrementally learning them. In this work, we follow existing related work \cite{OLN, LearningToDetectEveryThing} and  refer to "open-world" more generally as any setting or environment that contains OOD instances in the test set. Importantly, THPN would be very useful to incorporate into both open-set object detection models \textit{and} open-world object detection models.
  \end{itemize}
  
  \item Why do we not measure Average Precision?
  \begin{itemize}
      \item Computing precision for any class of object involves finding false positive predictions. In the case of class-agnostic open-set object detection, while we have labels for some of the OOD objects in the test images (e.g., the non-VOC COCO classes), this does not encapsulate \textit{all} OOD objects in the test images. For example, ``snowmobile" is not a COCO class, however there exists images with snowmobile instances in the validation set. It would be unfair to mark an accurate detection of an unlabeled snowmobile instance as a false positive. Therefore, we only consider recall. This is common practice in this field of study \cite{OLN, LearningToDetectEveryThing, KevinsPaper}. Furthermore, a proposal network's job is to maximize recall on all objects. In a full object detection model, there is a second stage which makes the final prediction refinement (e.g., R-CNN classification head).
  \end{itemize}

  \item How is AR-AUC computed?
  \begin{itemize}
      \item The most common metric for performance for proposal networks is Average Recall @ k detections per image (AR@k). The most common operating point to consider is AR@100, meaning the average recall when allowing 100 predictions per image. However, for some applications we may care about other operating points (e.g., k=10, k=1000, etc.). Therefore, a reasonable summary statistic to use is AUC introduced by Kim et al. \cite{OLN} which computes the area under the curve of AR@k for k=\{10, 30, 50, 100, 300, 500, 1000\}.
  \end{itemize}

  \item Does THPN's self-training require additional data that the baselines do not have access to?
  \begin{itemize}
      \item As mentioned in Section \ref{sec:experiments}, during training, not only do we assume access to \textit{only} ID labels of the ID classes in the training set, we also ensure that THPN is only ever exposed to images that contain at least one ID label during training and pseudo-label generation. Thus, our implementation of THPN does not use any unlabeled training images, just like any non-self-trained baseline. While our implementation of THPN does not use auxiliary images for self-training for the sake of fairness, it is certainly possible to do so. This can be seen as an additional degree of flexibility of our method (as self-training can seamlessly ingest unlabeled data). We save investigating this for future work.
  \end{itemize}
  
  \item Why is classification-based objectness beneficial for ID detection and localization-based objectness beneficial for OOD detection?
  \begin{itemize}
      \item Due to space constraints, we could not include all details about classification-based objectness and localization-based objectness in the main paper. However, we detail these methods in Appendix A. Essentially, the difference in behavior comes down to the fact that classification-based models rely on discriminative learning which forces the model to sample negative regions that do not contain an ID object. The problem is, these negative regions often contain OOD objects. Therefore, a classification-based model explicitly treats OOD objects as \textit{background}, meaning they are not detected during inference. Localization-based models instead frame objectness as the localization quality (e.g., centerness \cite{FCOS} or IoU \cite{IoUNet}) between a given region and any ground truth box. Critically, learning localization-quality is not discriminative, so it does not require explicitly sampling negative regions. Thus, this approach is inherently more likely to generalize to OOD objects \cite{OLN}. However, what is not shown in the OLN paper \cite{OLN} is the fact that this improved OOD recall comes at the cost of reduced ID recall. Our solution (THPN) offers a tunable, flexible model that can achieve the best of both worlds by learning and using both objectness representations with a weight $\lambda_{CLS}$.
  \end{itemize}
  
  \item Why do we not compare THPN to Open World Object Detection (OWOD) methods?
  \begin{itemize}
      \item While it may seem reasonable to compare THPN to the increasing number of OWOD models in the literature \cite{TowardsOpenWorldObjectDetection, RevisitingOWOD, OWDETR, UCOWOD, OWODDiscriminativeClassPrototype}, these models are for two very different tasks. OWOD models are full detection models that attempt to incrementally learn novel classes using a human-in-the-loop. THPN, on the other hand, is a class-agnostic proposal network. While THPN has the potential to greatly enhance the novel object detection capabilities as a component any OWOD model (i.e., by replacing RPN with THPN), we feel this investigation is beyond the scope of this paper, so we save it for future work.
  \end{itemize}

  \item Can THPN made to be dynamic at inference time instead of only tunable at training time?
  \begin{itemize}
      \item In our implementation of THPN, we assume that the $\lambda_{CLS}$ used during training is the same $\lambda_{CLS}$ used during inference. In this way, THPN is not necessarily meant to be tunable post-training. However, we also investigate what happens when we train with a moderate $\lambda_{CLS}=0.10$, and modulate $\lambda_{CLS}$ during inference. Results from this are in Appendix F. Overall, we find that while the results are not quite as good as when we modulate $\lambda_{CLS}$ during training, an inference-time-dynamic variant of THPN is still quite effective. Either variant outperforms all baselines.
  \end{itemize}

  \item What are the potential negative societal impacts?
  \begin{itemize}
      \item Powerful object detection algorithms enable a huge range of potential automated applications. While many of these are righteous technologies (e.g., autonomous vehicles, more effective search and rescue, robots for the disabled, etc.), there are of course several nefarious applications such as aggressive surveillance or weapon systems. Regardless, due to the ubiquity of these models in the public domain, we argue that making these models as robust and trustworthy as possible is overall a noble and important endeavor.
  \end{itemize}
  
\end{itemize}

%% file: tables/learning_free_baselines.tex
\begin{table}[h]
\centering
\resizebox{.35\columnwidth}{!}{

\begin{tabular}{lcc}
\toprule
                     & \multicolumn{2}{c}{ALL}       \\
Method               & AR10          & AR100         \\ \hline\hline \noalign{\vskip .6mm}
Geodesic \cite{GeodesicObjectProposals}             & 4.0           & 18.0          \\
Sel. Search \cite{SelectiveSearch}          & 5.2           & 16.3          \\
EdgeBoxes \cite{EdgeBoxes}            & 7.4           & 17.8          \\
MCG \cite{MCG}                  & 10.1          & 24.6          \\ \cdashline{1-3}  \noalign{\vskip .6mm}
\textbf{THPN ($\boldsymbol{\lambda_{CLS} = 0.50}$)}              & \textbf{41.0}          & \textbf{59.9} \\
\bottomrule
\end{tabular}

}
\vspace{-2mm}
\caption{ALL recall comparison with learning-free baselines on the COCO dataset. The THPN (trained on all COCO classes) significantly outperforms these approaches. Results for baselines are borrowed from \cite{DeepMask}.}
\label{tab:learning_free_baselines}

\end{table}

%% file: tables/closed_set.tex
\begin{table*}[h]
\centering
\resizebox{.7\linewidth}{!}{

\begin{tabular}{llcccccc}
\toprule
                       &               & \multicolumn{1}{l}{} & \multicolumn{1}{l}{} & \multicolumn{4}{c}{Test ALL} \\
Dataset                & Model         & Images               & Instances            & AUC  & AR10 & AR100 & AR1000 \\ \hline\hline \noalign{\vskip .6mm}
\multirow{9}{*}{COCO}  & Faster R-CNN  & 117k                 & 860k                 & 42.9 & 38.1 & 57.1  & 63.1   \\
                       & OLN           & 117k                 & 860k                 & 39.6 & 29.8 & 54.8  & 64.4   \\ \cdashline{2-8} \noalign{\vskip .6mm}
                       & THPN ($\lambda_{CLS}=0$)    & 117k                 & 860k                 & 39.6 & 29.8 & 54.8  & 64.4   \\
                       & THPN ($\lambda_{CLS}=0.10$) & 117k                 & 860k                 & 43.1 & 37.4 & 57.5  & 65.0   \\
                       & THPN ($\lambda_{CLS}=0.25$) & 117k                 & 860k                 & 44.7 & 40.4 & 59.1  & 65.6   \\
                       & \textbf{THPN ($\boldsymbol{\lambda_{CLS}=0.50}$)} & 117k                 & 860k                 & \textbf{45.3} & \textbf{41.0} & \textbf{59.9}  & \textbf{66.2}   \\
                       & THPN ($\lambda_{CLS}=0.75$) & 117k                 & 860k                 & 45.1 & 40.6 & 59.7  & 66.2   \\
                       & THPN ($\lambda_{CLS}=0.90$) & 117k                 & 860k                 & 44.8 & 40.1 & 59.3  & 66.2   \\
                       & THPN ($\lambda_{CLS}=1$)    & 117k                 & 860k                 & 44.7 & 39.9 & 59.2  & 66.0   \\ \hline \noalign{\vskip .6mm}
\multirow{9}{*}{Ships} & Faster R-CNN  & 2.2k                 & 10k                  & 50.2 & 51.5 & 63.1  & 68.4   \\
                       & OLN           & 2.2k                 & 10k                  & 51.4 & 51.5 & 65.7  & 70.3   \\ \cdashline{2-8} \noalign{\vskip .6mm}
                       & THPN ($\lambda_{CLS}=0$)    & 2.2k                 & 10k                  & 51.6 & 51.7 & \textbf{66.0}  & \textbf{70.3}   \\
                       & THPN ($\lambda_{CLS}=0.10$) & 2.2k                 & 10k                  & 51.6 & 52.7 & 65.6  & 69.6   \\
                       & \textbf{THPN ($\boldsymbol{\lambda_{CLS}=0.25}$)} & 2.2k                 & 10k                  & \textbf{51.8} & \textbf{53.1} & 65.8  & 69.7   \\
                       & THPN ($\lambda_{CLS}=0.50$) & 2.2k                 & 10k                  & 51.1 & 52.6 & 64.6  & 68.9   \\
                       & THPN ($\lambda_{CLS}=0.75$) & 2.2k                 & 10k                  & 51.4 & 52.8 & 65.2  & 69.4   \\
                       & THPN ($\lambda_{CLS}=0.90$) & 2.2k                 & 10k                  & 50.9 & 52.4 & 64.3  & 69.3   \\
                       & THPN ($\lambda_{CLS}=1$)    & 2.2k                 & 10k                  & 51.0 & 52.3 & 64.9  & 68.8  \\
\bottomrule
\end{tabular}

}
\vspace{-2mm}
\caption{Results on closed-set tasks. Here, we assume all instances of all classes of interest are labeled.}
\label{tab:closed_set}
\end{table*}

%% file: tables/augmentations.tex
\begin{table*}[h]
\centering
\resizebox{.7\linewidth}{!}{

\begin{tabular}{lcccc}
\toprule
Augmentation                          & Epochs & OOD-AUC       & ID-AUC        & ALL-AUC       \\ \hline\hline \noalign{\vskip .6mm}
None                                  & 8      & 24.8          & \textbf{44.8} & 35.5          \\
Crop \& Zoom                          & 8      & 25.5          & 43.0          & 34.8          \\
Discrete Rotate                       & 8      & 24.8          & 40.3          & 33.0          \\
Random Affine                         & 8      & 23.5          & 42.3          & 33.6          \\
Photometric Distortion                & 8      & 25.0          & 44.4          & 35.4          \\ 
Crop \& Zoom + Photometric Distortion & 8      & 25.5          & 42.0          & 34.2          \\  \cdashline{1-5} \noalign{\vskip .6mm}
None                                  & 16     & 24.8          & 44.8          & 35.5          \\
Crop \& Zoom                          & 16     & \textbf{26.1} & 44.2          & \textbf{35.8} \\
Discrete Rotate                       & 16     & 25.4          & 41.8          & 34.0          \\
Random Affine                         & 16     & 24.0          & 43.7          & 34.5          \\
Crop \& Zoom + Photometric Distortion & 16     & 26.0          & 43.7          & 35.4         \\
\bottomrule
\end{tabular}

}
\vspace{-2mm}
\caption{Effect of augmentation and training schedule on an OLN model trained on the VOC split.}
\label{tab:augmentations}

\end{table*}

%% file: tables/ships_challenge_full.tex
\begin{table*}[t]
\centering
\resizebox{.95\linewidth}{!}{

\begin{tabular}{llc|cccc|cccc|cccc}
\toprule
                          &                                      & Images /    & \multicolumn{4}{c|}{OOD}                                      & \multicolumn{4}{c|}{ID}                                       & \multicolumn{4}{c}{ALL}                                       \\
Split                     & Model                                & Instances   & AUC           & AR10          & AR100         & AR1k          & AUC           & AR10          & AR100         & AR1k          & AUC           & AR10          & AR100         & AR1k          \\ \hline\hline \noalign{\vskip .6mm}
\multirow{6}{*}{Military}  & Faster R-CNN                         & 1.5k / 4.7k & 11.8          & 11.8          & 12.2          & 24.3          & 65.6          & 75.6          & 79.4          & 81.3          & 33.3          & 37.1          & 39.1          & 47.1          \\
                          & OLN                                  & 1.5k / 4.7k & 23.0          & 23.4          & 28.1          & 35.8          & 69.0          & 79.2          & 84.4          & 85.4          & 41.3          & 45.2          & 50.6          & 55.6          \\ \cdashline{2-15} \noalign{\vskip .6mm}
                          & \textbf{THPN ($\boldsymbol{\lambda_{CLS}=0}$)} & 1.5k / 6.1k & \textbf{29.0} & \textbf{28.4} & \textbf{37.1} & \textbf{40.7} & 68.8          & 78.3          & 84.2          & 85.1          & \textbf{44.7} & \textbf{47.7} & \textbf{56.0} & \textbf{58.5} \\
                          & THPN ($\lambda_{CLS}=0.10$)       & 1.5k / 6.1k & 26.6          & 27.0          & 33.6          & 37.8          & \textbf{69.3} & \textbf{79.4}          & \textbf{84.7} & \textbf{85.4} & 43.6          & 47.5          & 54.1          & 56.9          \\
                          & THPN ($\lambda_{CLS}=0.25$)       & 1.5k / 6.1k & 22.4          & 23.1          & 27.3          & 33.8          & 69.0          & 79.3 & 84.1          & 84.9          & 40.9          & 45.2          & 50.0          & 54.3          \\
                          & THPN ($\lambda_{CLS}=0.50$)       & 1.5k / 6.1k & 16.1          & 16.8          & 19.2          & 25.8          & 68.3          & 78.7          & 83.1          & 84.4          & 36.9          & 41.3          & 44.7          & 49.2          \\ \hline
\multirow{6}{*}{Civilian} & Faster R-CNN                         & 1.0k / 4.4k & 16.0          & 16.8          & 17.8          & 29.1          & 33.4          & 29.5          & 44.3          & 51.2          & 24.6          & 22.6          & 31.0          & 40.1          \\
                          & OLN                                  & 1.0k / 4.4k & 38.7          & 38.4          & 49.2          & 56.6          & 35.6          & 30.7          & 47.9          & 54.3          & 36.9          & 33.8          & 48.5          & 55.5          \\ \cdashline{2-15} \noalign{\vskip .6mm}
                          & \textbf{THPN ($\boldsymbol{\lambda_{CLS}=0}$)} & 1.0k / 5.7k & \textbf{49.8} & \textbf{50.4} & \textbf{63.5} & \textbf{68.3} & \textbf{36.5} & 30.3          & \textbf{49.1} & \textbf{56.7} & \textbf{42.8} & \textbf{39.3} & \textbf{56.2} & \textbf{62.5} \\
                          & THPN ($\lambda_{CLS}=0.10$)       & 1.0k / 5.7k & 46.3          & 47.2          & 58.4          & 64.2          & 36.4          & 31.4          & 48.5          & 55.0          & 41.0          & 38.2          & 53.4          & 59.6          \\
                          & THPN ($\lambda_{CLS}=0.25$)       & 1.0k / 5.7k & 33.9          & 34.4          & 41.6          & 52.5          & 35.8          & \textbf{31.5} & 47.6          & 54.0          & 34.6          & 32.0          & 44.5          & 53.3          \\
                          & THPN ($\lambda_{CLS}=0.50$)       & 1.0k / 5.7k & 24.1          & 22.8          & 28.7          & 42.8          & 35.5          & 31.1          & 46.9          & 54.1          & 29.5          & 26.2          & 37.7          & 48.5       
                        \\
\bottomrule
\end{tabular}

}
\vspace{-2mm}
\caption{Full results on the \textit{ships} challenge.}
\label{tab:ships_challenge_full}
\end{table*}

%% file: tables/tl10_training_class_diversity_challenge.tex
\begin{table*}[t]
\centering
\resizebox{0.95\linewidth}{!}{

\begin{tabular}{llc|cccc|cccc|cccc}
\toprule
                        &                                         & Images /    & \multicolumn{4}{c|}{OOD}                                      & \multicolumn{4}{c|}{ID}                                       & \multicolumn{4}{c}{ALL}                                       \\
Split                   & Model                                   & Instances   & AUC           & AR10          & AR100         & AR1k        & AUC           & AR10          & AR100         & AR1k        & AUC           & AR10          & AR100         & AR1k        \\ \hline\hline \noalign{\vskip .6mm}
\multirow{6}{*}{COCO40} & Faster R-CNN                            & 104k / 623k & 26.6          & 17.5          & 36.0          & 51.4          & 44.4          & 41.4          & 58.3          & 63.2          & 39.0          & 33.8          & 51.7          & 60.0          \\
                        & OLN                                     & 104k / 623k & 33.1          & 25.8          & 44.8          & 54.6          & 42.1          & 34.6          & 57.2          & 65.0          & 38.9          & 30.5          & 53.3          & 62.2          \\ \cdashline{2-15} \noalign{\vskip .6mm}
                        & THPN ($\lambda_{CLS}=0$)             & 104k / 810k & 34.1          & 28.0          & 45.6          & 55.2  & 41.6          & 34.6          & 56.2          & 64.0          & 38.9          & 31.1          & 52.8          & 61.6          \\
                        & THPN ($\lambda_{CLS}=0.10$)          & 104k / 810k & \textbf{34.8} & \textbf{29.8} & \textbf{46.0} & 55.3          & 44.0          & 39.6          & 58.1          & 64.6          & 40.7          & 35.1          & 54.3          & 62.0          \\
                        & \textbf{THPN ($\boldsymbol{\lambda_{CLS}=0.25}$)} & 104k / 810k & 34.5          & 29.2          & 45.6          & \textbf{55.4}          & 45.0          & 41.5          & 59.0          & \textbf{65.0}          & \textbf{41.5} & \textbf{36.5} & \textbf{54.9} & \textbf{62.4}          \\
                        & THPN ($\lambda_{CLS}=0.50$)          & 104k / 810k & 33.6          & 28.0          & 44.5          & 55.3          & \textbf{45.1} &  \textbf{41.7} &   \textbf{60.7} &         64.7 & 34.4 & 29.9          & 44.8          & 55.1         \\ \hline \noalign{\vskip .6mm}
\multirow{6}{*}{VOC}    & Faster R-CNN                            & 95k / 493k  & 19.3          & 11.6          & 25.1          & 42.4          & 46.7          & 45.1          & 60.7          & 64.7          & 34.4          & 29.9          & 44.8          & 55.1          \\
                        & OLN                                     & 95k / 493k  & 24.8          & 18.4          & 33.2          & 45.0          & 44.8          & 40.1          & 59.3          & 66.1          & 35.5          & 29.1          & 47.5          & 56.9          \\ \cdashline{2-15} \noalign{\vskip .6mm}
                        & THPN ($\lambda_{CLS}=0$)             & 95k / 641k  & 27.7 & 21.3          & 36.9 & 48.0 & 44.8          & 39.9          & 59.5          & 66.0          & 36.7          & 30.1          & 49.3          & 58.3 \\
                        & THPN ($\lambda_{CLS}=0.10$) & 95k / 641k  & \textbf{27.9}          & \textbf{22.0} & \textbf{37.1}          & \textbf{48.0}          & 46.8          & 44.2          & 60.9          & 66.5          & 38.0 & 32.9          & 50.2 & 58.5          \\
                        & \textbf{THPN ($\boldsymbol{\lambda_{CLS}=0.25}$)}          & 95k / 641k  & 27.5          & 21.4          & 36.6          & 48.0          & 47.6          & 45.8          & 61.8          & 66.8          & \textbf{38.4}          & \textbf{33.7} & \textbf{50.5}          & 58.8          \\
                        & THPN ($\lambda_{CLS}=0.50$)          & 95k / 641k  & 26.4          & 19.8          & 35.0          & 47.8          & \textbf{47.8} & \textbf{46.1} & \textbf{62.1} & \textbf{67.1} & 38.0          & 33.5          & 50.0          & \textbf{58.9}          \\ \hline \noalign{\vskip .6mm}
\multirow{6}{*}{VOC5}   & Faster R-CNN                            & 74k / 357k  & 16.3          & 9.8           & 20.7          & 38.1          & 48.1          & 47.6          & 62.2          & 65.6          & 29.1          & 24.8          & 37.4          & 49.6          \\
                        & OLN                                     & 74k / 357k  & 20.3          & 14.1          & 26.9          & 40.1          & 47.6          & 45.2          & 61.7          & 67.8          & 31.0          & 25.7          & 40.8          & 51.6          \\ \cdashline{2-15} \noalign{\vskip .6mm}
                        & THPN ($\lambda_{CLS}=0$)    & 74k / 465k  & 23.5 & 17.1 & 31.3 & 43.9 & 46.8          & 43.9          & 61.1          & 67.0          & 32.6  & 26.7          & 43.2 & 53.7 \\
                        & THPN ($\lambda_{CLS}=0.10$)          & 74k / 465k  & \textbf{23.7}          & \textbf{17.6}          & \textbf{31.5}          & \textbf{43.9}          & 48.3          & 47.1          & 62.4          & 67.4          & 33.3          & 28.4 & 43.8          & 53.7          \\
                        & \textbf{THPN ($\boldsymbol{\lambda_{CLS}=0.25}$)}          & 74k / 465k  & 23.6          & 17.4          & 31.1          & 43.9          & 49.0          & 48.4          & 63.1          & 67.7          & \textbf{33.5}          & \textbf{28.9}          & \textbf{43.9}          & \textbf{53.8}          \\
                        & THPN ($\lambda_{CLS}=0.50$)          & 74k / 465k  & 22.7          & 16.1          & 30.0          & 43.9          & \textbf{49.3} & \textbf{48.7} & \textbf{63.5} & \textbf{67.8} & 33.2          & 28.5          & 43.4          & 53.8          \\ \hline \noalign{\vskip .6mm}
\multirow{6}{*}{Animal} & Faster R-CNN                            & 24k / 63k   & 11.5          & 6.0           & 13.5          & 31.3          & 53.9          & 58.9          & 67.1          & 69.4          & 14.6          & 9.8           & 17.5          & 34.1          \\
                        & OLN                                     & 24k / 63k   & 13.3          & 8.2           & 16.4          & 31.5          & 55.8          & 59.7          & 69.7          & 73.2          & 16.4          & 11.9          & 20.3          & 34.6          \\ \cdashline{2-15} \noalign{\vskip .6mm}
                        & THPN ($\lambda_{CLS}=0$)    & 24k / 81k   & 16.9 & 9.9          & 22.9 & \textbf{36.7} & 55.5          & 59.0          & 69.7          & 73.0          & 19.7  & 13.4          & 26.3 & 39.3 \\
                        & THPN ($\lambda_{CLS}=0.10$)          & 24k / 81k   & \textbf{17.0}          & \textbf{10.4} & \textbf{22.9}          & 36.6          & 56.1          & 60.6          & 69.9          & 73.0          & 19.8          & 13.9 & 26.3          & 39.3          \\
                        & \textbf{THPN ($\boldsymbol{\lambda_{CLS}=0.25}$)}          & 24k / 81k   & 17.0          & 10.3          & 22.8          & 36.6          & 56.6          & 61.5          & 70.5          & 73.2          & \textbf{19.8}          & \textbf{14.0}          & \textbf{26.3}          & \textbf{39.3}          \\
                        & THPN ($\lambda_{CLS}=0.50$)          & 24k / 81k   & 15.9          & 9.3           & 20.7          & 36.6          & \textbf{56.8} & \textbf{61.9} & \textbf{70.8} & \textbf{73.3} & 18.9          & 13.1          & 24.4          & 39.3    
                        \\
\bottomrule
\end{tabular}

}
\vspace{-2mm}
\caption{Results on the \textit{training class diversity} challenge when using a THPN trained with $\lambda_{CLS}=0.10$. The listed $\lambda_{CLS}$ in the table is the value used during inference-time.}
\label{tab:tl10_training_class_diversity_challenge}
\vspace{-3mm}
\end{table*}

%% file: tables/tl10_semi_supervised_challenge.tex
\begin{table*}[t]
\centering
\resizebox{0.95\linewidth}{!}{

\begin{tabular}{llc|cccc|cccc|cccc}
\toprule
                            &                                         & Images /   & \multicolumn{4}{c|}{OOD}                                      & \multicolumn{4}{c|}{ID}                                       & \multicolumn{4}{c}{ALL}                                       \\
Split                       & Model                                   & Instances  & AUC           & AR10          & AR100         & AR1k          & AUC           & AR10          & AR100         & AR1k          & AUC           & AR10          & AR100         & AR1k          \\ \hline\hline \noalign{\vskip .6mm}
\multirow{6}{*}{VOC (50\%)} & Faster R-CNN                            & 75k / 246k & 18.7          & 11.7          & 24.1          & 40.9          & 44.8          & 42.7          & 58.5          & 63.1          & 33.1          & 28.5          & 43.2          & 53.6          \\
                            & OLN                                     & 75k / 246k & 23.8          & 17.7          & 31.7          & 43.8          & 44.4          & 39.5          & 58.8          & 65.7          & 34.9          & 28.5          & 46.7          & 56.3          \\ \cdashline{2-15} \noalign{\vskip .6mm}
                            & THPN ($\lambda_{CLS}=0$)             & 75k / 320k & 25.7 & 19.4 & \textbf{34.2} & \textbf{45.9} & 44.7          & 40.9          & 58.6          & 65.4          & 35.9          & 30.1          & 47.6 & 57.1 \\
                            & THPN ($\lambda_{CLS}=0.10$) & 75k / 320k & \textbf{25.7}          & \textbf{19.6}          & 34.0          & 45.9          & 46.1          & 44.1          & 59.7          & 65.7          & 36.7 & 32.0 & 48.1          & 57.2          \\
                            & \textbf{THPN ($\boldsymbol{\lambda_{CLS}=0.25}$)}          & 75k / 320k & 25.1          & 18.5         & 33.3          & 45.8          & 46.8          & \textbf{45.3} & 60.5          & 66.0          & \textbf{36.9}          & \textbf{32.4}          & \textbf{48.3}          & \textbf{57.4}          \\
                            & THPN ($\lambda_{CLS}=0.50$)          & 75k / 320k & 23.7          & 16.3          & 31.4          & 45.7          & \textbf{46.8} & 44.9          & \textbf{60.8} & \textbf{66.2} & 36.3          & 31.5          & 47.6          & 57.4          \\ \hline \noalign{\vskip .6mm}
\multirow{6}{*}{VOC (25\%)} & Faster R-CNN                            & 56k / 123k & 17.9          & 11.2          & 22.9          & 39.2          & 42.7          & 40.1          & 55.8          & 60.9          & 31.6          & 27.0          & 41.1          & 51.6          \\
                            & OLN                                     & 56k / 123k & 21.9          & 16.6          & 28.8          & 40.7          & 43.2          & 38.3          & 57.1          & 64.1          & 33.4          & 27.5          & 44.5          & 54.0          \\ \cdashline{2-15} \noalign{\vskip .6mm}
                            & THPN ($\lambda_{CLS}=0$)             & 56k / 160k & \textbf{24.3} & \textbf{17.9} & \textbf{32.3} & \textbf{44.8} & 43.6          & 39.6          & 57.2          & 64.1          & 34.6          & 28.8          & 46.0 & 55.8 \\
                            & THPN ($\lambda_{CLS}=0.10$) & 56k / 160k & 24.2          & 17.9          & 32.1          & 44.8          & 44.7          & 42.4          & 58.0          & 64.2          & 35.3 & 30.5          & 46.4          & 55.9          \\
                            & \textbf{THPN ($\boldsymbol{\lambda_{CLS}=0.25}$)}           & 56k / 160k & 23.7          & 17.0          & 31.4          & 44.7          & 45.2          & \textbf{43.3} & 58.5          & 64.3          & \textbf{35.4}          & \textbf{30.7} & \textbf{46.4}          & \textbf{56.0}          \\
                            & THPN ($\lambda_{CLS}=0.50$)          & 56k / 160k & 22.6          & 15.2          & 29.9          & 44.6          & \textbf{45.3} & 43.1          & \textbf{58.8} & \textbf{64.5} & 35.0          & 30.0          & 45.8          & 56.0          \\ \hline \noalign{\vskip .6mm}
\multirow{6}{*}{VOC (10\%)} & Faster R-CNN                            & 33k / 49k  & 16.2          & 10.4          & 20.5          & 35.8          & 39.5          & 36.2          & 51.8          & 57.8          & 29.1          & 24.5          & 37.9          & 48.4          \\
                            & OLN                                     & 33k / 49k  & 19.8          & 15.2          & 25.7          & 37.3          & 40.8          & 36.3          & 53.6          & 61.0          & 31.3          & 26.0          & 41.3          & 50.8          \\ \cdashline{2-15} \noalign{\vskip .6mm}
                            & THPN ($\lambda_{CLS}=0$)             & 33k / 64k  & 22.9 & 16.9 & \textbf{30.3} & \textbf{42.4} & 41.5          & 37.3          & 54.5          & 61.8          & 33.0          & 27.2          & 43.7 & 53.5 \\
                            & THPN ($\lambda_{CLS}=0.10$) & 33k / 64k  & \textbf{23.0}          & \textbf{17.2}          & 30.2          & 42.4          & 42.3          & 39.5          & 55.0          & 61.9          & 33.5 & 28.6 & 43.9          & 53.6          \\
                            & \textbf{THPN ($\boldsymbol{\lambda_{CLS}=0.25}$)}           & 33k / 64k  & 22.6          & 16.4         & 29.9          & 42.4          & 42.8          & \textbf{40.4} & 55.5          & 62.1          & \textbf{33.6}          & \textbf{28.9}          & \textbf{44.1}          & 53.7          \\
                            & THPN ($\lambda_{CLS}=0.50$)          & 33k / 64k  & 21.9          & 15.0          & 29.1          & 42.5          & \textbf{43.0} & 40.3 & \textbf{55.9}          & \textbf{62.2}  & 33.4          & 28.4          & 44.0          & \textbf{53.8}       
                        \\
\bottomrule
\end{tabular}

}
\vspace{-2mm}
\caption{Results on the \textit{semi-supervised} challenge when using a THPN trained with $\lambda_{CLS}=0.10$. The listed $\lambda_{CLS}$ in the table is the value used during inference-time.}
\label{tab:tl10_semi_supervised_challenge}
\vspace{-2mm}
\end{table*}

%% file: tables/coco_splits.tex
\begin{table*}[h]
\centering
\resizebox{0.35\linewidth}{!}{
\begin{tabular}{lccccc}
\toprule
COCO Class      & Super Category & COCO40       & VOC          & VOC5         & Animal       \\ \hline\hline \noalign{\vskip .6mm}
person          & person         & $\checkmark$ & $\checkmark$ & $\checkmark$ &              \\
\cdashline{1-6} \noalign{\vskip .6mm}
bicycle         & vehicle        & $\checkmark$ & $\checkmark$ & $\checkmark$ &              \\
car             & vehicle        & $\checkmark$ & $\checkmark$ & $\checkmark$ &              \\
motorcycle      & vehicle        & $\checkmark$ & $\checkmark$ &              &              \\
airplane        & vehicle        & $\checkmark$ & $\checkmark$ &              &              \\
bus             & vehicle        & $\checkmark$ & $\checkmark$ &              &              \\
train           & vehicle        & $\checkmark$ & $\checkmark$ &              &              \\
truck           & vehicle        &              &              &              &              \\
boat            & vehicle        & $\checkmark$ & $\checkmark$ &              &              \\
\cdashline{1-6} \noalign{\vskip .6mm}
traffic light   & outdoor        & $\checkmark$ &              &              &              \\
fire hydrant    & outdoor        &              &              &              &              \\
stop sign       & outdoor        & $\checkmark$ &              &              &              \\
parking meter   & outdoor        &              &              &              &              \\
bench           & outdoor        & $\checkmark$ &              &              &              \\
\cdashline{1-6} \noalign{\vskip .6mm}
bird            & animal         & $\checkmark$ & $\checkmark$ &              & $\checkmark$ \\
cat             & animal         & $\checkmark$ & $\checkmark$ &              & $\checkmark$ \\
dog             & animal         & $\checkmark$ & $\checkmark$ & $\checkmark$ & $\checkmark$ \\
horse           & animal         & $\checkmark$ & $\checkmark$ &              & $\checkmark$ \\
sheep           & animal         & $\checkmark$ & $\checkmark$ &              & $\checkmark$ \\
cow             & animal         & $\checkmark$ & $\checkmark$ &              & $\checkmark$ \\
elephant        & animal         &              &              &              & $\checkmark$ \\
bear            & animal         &              &              &              & $\checkmark$ \\
zebra           & animal         &              &              &              & $\checkmark$ \\
giraffe         & animal         &              &              &              & $\checkmark$ \\
\cdashline{1-6} \noalign{\vskip .6mm}
backpack        & accessory      & $\checkmark$ &              &              &              \\
umbrella        & accessory      &              &              &              &              \\
handbag         & accessory      & $\checkmark$ &              &              &              \\
tie             & accessory      &              &              &              &              \\
suitcase        & accessory      &              &              &              &              \\
\cdashline{1-6} \noalign{\vskip .6mm}
frisbee         & sports         &              &              &              &              \\
skis            & sports         & $\checkmark$ &              &              &              \\
snowboard       & sports         &              &              &              &              \\
sports ball     & sports         & $\checkmark$ &              &              &              \\
kite            & sports         &              &              &              &              \\
baseball bat    & sports         &              &              &              &              \\
baseball glove  & sports         &              &              &              &              \\
skateboard      & sports         & $\checkmark$ &              &              &              \\
surfboard       & sports         & $\checkmark$ &              &              &              \\
tennis racket   & sports         &              &              &              &              \\
\cdashline{1-6} \noalign{\vskip .6mm}
bottle          & kitchen        & $\checkmark$ & $\checkmark$ &              &              \\
wine glass      & kitchen        &              &              &              &              \\
cup             & kitchen        &              &              &              &              \\
fork            & kitchen        & $\checkmark$ &              &              &              \\
knife           & kitchen        &              &              &              &              \\
spoon           & kitchen        &              &              &              &              \\
bowl            & kitchen        & $\checkmark$ &              &              &              \\
\cdashline{1-6} \noalign{\vskip .6mm}
banana          & food           &              &              &              &              \\
apple           & food           & $\checkmark$ &              &              &              \\
sandwich        & food           &              &              &              &              \\
orange          & food           &              &              &              &              \\
broccoli        & food           &              &              &              &              \\
carrot          & food           &              &              &              &              \\
hot dog         & food           &              &              &              &              \\
pizza           & food           & $\checkmark$ &              &              &              \\
donut           & food           &              &              &              &              \\
cake            & food           &              &              &              &              \\
\cdashline{1-6} \noalign{\vskip .6mm}
chair           & furniture      & $\checkmark$ & $\checkmark$ & $\checkmark$ &              \\
couch           & furniture      & $\checkmark$ & $\checkmark$ &              &              \\
potted plant    & furniture      & $\checkmark$ & $\checkmark$ &              &              \\
bed             & furniture      &              &              &              &              \\
dining table    & furniture      & $\checkmark$ & $\checkmark$ &              &              \\
toilet          & furniture      & $\checkmark$ &              &              &              \\
\cdashline{1-6} \noalign{\vskip .6mm}
tv              & electronic     & $\checkmark$ & $\checkmark$ &              &              \\
laptop          & electronic     & $\checkmark$ &              &              &              \\
mouse           & electronic     &              &              &              &              \\
remote          & electronic     & $\checkmark$ &              &              &              \\
keyboard        & electronic     &              &              &              &              \\
cell phone      & electronic     &              &              &              &              \\
\cdashline{1-6} \noalign{\vskip .6mm}
microwave       & appliance      &              &              &              &              \\
oven            & appliance      & $\checkmark$ &              &              &              \\
toaster         & appliance      &              &              &              &              \\
sink            & appliance      & $\checkmark$ &              &              &              \\
refrigerator    & appliance      & $\checkmark$ &              &              &              \\
\cdashline{1-6} \noalign{\vskip .6mm}
book            & indoor         &              &              &              &              \\
clock           & indoor         &              &              &              &              \\
vase            & indoor         &              &              &              &              \\
scissors        & indoor         &              &              &              &              \\
teddy bear      & indoor         &              &              &              &              \\
hair drier      & indoor         &              &              &              &              \\
toothbrush      & indoor         & $\checkmark$ &              &              &              \\ \hline \noalign{\vskip .6mm}
\multicolumn{2}{l}{\% Train Set} & 72.5         & 57.3         & 41.6         & 7.3          \\
\multicolumn{2}{l}{\% Test Set}  & 72.2         & 57.0           & 41.5         & 7.3   
\\
\bottomrule
\end{tabular}
}

\caption{Training splits for the \textit{training class diversity} challenge.}
\label{tab:coco_splits}

\end{table*}